\documentclass[letterpaper]{article} 
\usepackage{aaai2026}  
\usepackage{times}  
\usepackage{helvet}  
\usepackage{courier}  
\usepackage[hyphens]{url}  
\usepackage{graphicx} 
\usepackage{natbib}  
\usepackage{caption} 
\usepackage{algorithm}
\usepackage{algorithmic}

\usepackage{newfloat}
\usepackage{listings}
\DeclareCaptionStyle{ruled}{labelfont=normalfont,labelsep=colon,strut=off} 
\floatstyle{ruled}
\newfloat{listing}{tb}{lst}{}
\floatname{listing}{Listing}
\usepackage{xcolor}

\title{Don't Stop the Multi-Party! \\ On Generating Synthetic  Written  Multi-Party Conversations \\ with Constraints}
\author{
    Nicol\`o Penzo\textsuperscript{\rm 1,2}, Marco Guerini\textsuperscript{\rm 1}, Bruno Lepri\textsuperscript{\rm 1}, Goran Glava\v{s}\textsuperscript{\rm 3}, Sara Tonelli\textsuperscript{\rm 1}
}
\affiliations{
    \textsuperscript{\rm 1}Fondazione Bruno Kessler, Italy\\
    \textsuperscript{\rm 2}University of Trento, Italy\\
    \textsuperscript{\rm 3}Center for Artificial Intelligence and Data Science, University of W\"urzburg, Germany
    
    \{npenzo, guerini, lepri, satonelli\}@fbk.eu, goran.glavas@uni-wuerzburg.de
}

\begin{document}
\maketitle

\begin{abstract}

 Written Multi-Party Conversations (WMPCs)  are widely studied across disciplines, with social media as a primary data source due to their accessibility. However, these datasets raise privacy concerns and often reflect  platform-specific properties.
For example, interactions between speakers may be limited due to rigid platform structures (e.g., threads, tree-like discussions), which yield overly simplistic interaction patterns (e.g., one-to-one ``reply-to'' links). This work explores the feasibility of generating synthetic WMPCs with instruction-tuned Large Language Models (LLMs) by providing deterministic constraints such as dialogue structure and participants’ stance. We investigate two complementary \textit{strategies} of leveraging LLMs in this context: \textsc{(i.)} \textit{LLMs as WMPC generators}, where we task the LLM to generate a whole WMPC at once and \textsc{(ii.)} \textit{LLMs as WMPC parties}, where the LLM generates one turn of the conversation at a time (made of speaker, addressee and message), provided the conversation history. We next introduce an analytical framework to evaluate compliance with the constraints, content quality, and interaction complexity for both strategies. Finally, we assess the level of obtained WMPCs via human and LLM-as-a-judge evaluations. We find stark differences among LLMs, with only some being able to generate high-quality WMPCs. We also find that turn-by-turn generation yields better conformance to constraints and higher linguistic variability than generating WMPCs in one pass. Nonetheless, our structural and qualitative evaluation indicates that both generation strategies can yield high-quality WMPCs. 

\end{abstract}

\begin{links}
    \link{Code \& Dataset}{https://github.com/dhfbk/Constrained-SyntheticMPC}
    \link{Published version in AAAI2026}{https://doi.org/10.1609/aaai.v40i39.40548}
\end{links}

\section{Introduction}\label{sec:intro}

Multi-Party Conversations (MPCs), i.e., conversations involving more than two participants \cite{branigan2006perspectives}, have been studied across multiple disciplines. Research in conversational analysis and linguistics has focused on modeling interaction dynamics \cite{sacks1978simplest, wilson1984models},  identifying participant roles \cite{malouf1995towards}, or mapping emergent structural patterns in discourse \cite{10.1353/sof.2003.0055}. These studies highlight both complexity and diversity of real-world MPCs, where factors like turn-taking, speaker alignment, and social context shape the flow of conversation.

The collection of MPC data has, however, strongly shifted from in-person and online meetings to social media platforms \cite{mahajan-shaikh-2021-need}, where large-scale data is more accessible. However, this shift has introduced several confounding factors. Social media platforms often enforce a one-to-one reply structure, overlooking implicit addressees and simplifying interaction dynamics; in natural conversations, in contrast, a turn is often directed to multiple participants and the conversational structure is more dynamic.  Furthermore, the asynchronous nature of social media eliminates overlapping turns, resulting in a well-defined sequence of utterances. For these reasons, we refer to conversations drawn from such platforms as Written Multi-Party Conversations (WMPCs.) 
As a result, WMPC corpora derived from social media platforms often lack structural diversity, which severely limits their utility in analyzing real-world conversational phenomena \cite{weiEtAl2023}. 
This, in turn, affects generation capabilities of current Large Language Models (LLMs). For LLMs, trained on conversations from social media and predominantly used in two-party interactions (i.e. human-assistant use-cases), WMPCs represent a distributional shift, resulting in their underwhelming performance in natural WMPC contexts \cite{tan-etal-2023-chatgpt, penzo-etal-2024-llms}. The next generation of LLMs is, however, expected to engage in MPCs and excel in tasks like identifying the appropriate speaker to respond to \cite{weiEtAl2023}, summarizing meetings \cite{kirstein-etal-2024-tell} or even managing multi-agent scenarios \cite{wu2023autogenenablingnextgenllm}. Recent studies have explored their performance in social contexts \cite{10.1162/coli_a_00502, chang2024llmsgeneratestructurallyrealistic}, emphasizing the need for large, representative datasets to train this novel generation of LLMs and to ensure robustness across diverse and less frequent interaction patterns \cite{lee2024towards}. 

One potential remedy for the lack of structural diversity in WMPCs derived from social media data is to synthesize WMPCs, by \textit{explicitly constraining LLMs} to generate WMPCs with specific characteristics, such as number of messages, number of speakers, speakers' stance, output format or interaction rules. To reflect real-world conversational complexity, generated MPCs should include varied conversations, encompass different interaction patterns and topics as well as provide rich speaker-addressee relationships, e.g., multi-addressee interactions. 

In this paper, we propose generating synthetic WMPCs using LLMs guided by constraints related to the above capabilities. We explore two generation strategies: \textsc{(i.)} One-Long (OL) generation, where the LLM produces an entire WMPC in a single step, and \textsc{(ii.)} Turn-by-Turn (TT) generation, which constructs the conversation sequentially, one turn at a time. A comparison of resulting WMPCs from both strategies highlights the (potential) discrepancies between 
\textsc{(i.)} how LLMs cast entire WMPCs to look human-like (OL) and \textsc{(ii.)} how they behave as participants in a WMPC (TT).
We propose a novel evaluation framework that combines several quantitative and qualitative dimensions of generated WMPCs, focusing on the extent of LLMs' compliance to provided content and structural constraints. 
We address the following three key research questions:

\vspace{0.8mm}
\noindent \textbf{RQ(1)}: Can LLMs be leveraged to generate large synthetic WMPC datasets while maintaining compliance with predefined constraints on dialogue structure and participants' stance?

\vspace{0.8mm}
\noindent \textbf{RQ(2)}: Which generation strategy (One-Long vs. Turn-by-Turn) produces higher-quality WMPCs?

\vspace{0.8mm}
\noindent\textbf{RQ(3)}: How can we effectively evaluate the variety and quality of the generated WMPCs?

\vspace{1mm}
We test four popular LLMs and identify Llama3.1 \cite{dubey2024llama3herdmodels} and Qwen2.5 \cite{yang2024qwen2technicalreport} as the best LLMs for complying the most with constraints. TT seems to generate more constraint-compliant WMPCs than OL. Moreover, the WMPCs produced by TT exhibit greater lexical variability and semantic coherence. The generated WMPCs also present a higher structural complexity than a widely-used corpus of ``real'' conversations \cite{ouchi-tsuboi-2016-addressee}. Finally, a qualitative evaluation shows that both TT and OL can produce high-quality WMPCs, rendering the choice of the LLM more important than the choice of generation strategy.

\section{Related Work}
\citet{mahajan-shaikh-2021-need} categorize MPC corpora into three types: Spoken Unscripted, Spoken Scripted, and Written. In this section, we discuss unscripted MPCs (both Spoken and Written). 

Corpora containing transcriptions of spoken unscripted MPCs typically rely on in-person meetings, e.g.  AMI \cite{10.1007/11677482_3} and ICSI \cite{1198793}. Non-verbal cues, the physical environment and overlapping turns are elements that significantly shape such interaction dynamics, while these factors are typically absent in written interactions. 
Moreover, these datasets lack addressee information, making them unsuitable for our intended analyses.

Written MPCs (WMPCs), on the other hand, are characteristic of online platforms where conversations unfold asynchronously without overlapping.
While social media allow for rapid collection of large-scale WMPCs, these datasets often come with incomplete interaction metadata.
Many datasets record only explicit reply-to relationships, neglecting implicit addressees and richer conversational dynamics \cite{ouchi-tsuboi-2016-addressee, zhang-etal-2018-conversations, chang-danescu-niculescu-mizil-2019-trouble}. \citet{weiEtAl2023} point out that well-known WMPC corpora \cite{ritter-etal-2010-unsupervised, baumgartner2020pushshiftredditdataset, lowe-etal-2015-ubuntu} are useful for response generation, but not for more interactive tasks. Only most recent efforts focus on capturing conversational dynamics, i.e., going beyond text content \cite{penzo-etal-2024-putting, hua-etal-2024-get}.
Among these datasets, the Ubuntu IRC corpus \cite{ouchi-tsuboi-2016-addressee} is the only dataset that we can use as a comparison for this work. 
Indeed, it is possible to retain from the initial set of $700\,000$ WMPCs only those with the same number of messages, number of users and structural constraints as in our synthetic WMPCs, obtaining a set of conversations with a size comparable to our synthetic datasets (details in Section \ref{subsec:global-analysis-expl}).

Structural analyses of social communication networks have primarily focused on interaction patterns across multiple conversations \cite{coletto2017motif, 10.1145/3178876.3186139, felmlee2021dyads}. This confirms the relevance of such structures in studying conversation dynamics. However, our focus is on interactions emerging within a single conversation rather than across multiple discussions, applying the same structural analysis techniques. 

To the best of our knowledge, the only existing attempt at generating synthetic WMPCs was made by \citet{chen-etal-2023-places}. However, their work primarily focused on conversations involving at most three participants, limiting the complexity of interactions. In contrast, our study explores the generation of WMPCs with four or more participants, leading to more elaborate  
discussion dynamics. While this increased complexity allows for richer conversational structures, it also introduces a higher likelihood of generation errors, necessitating a rigorous evaluation process to assess the quality and consistency of the generated dialogues.

\section{Synthetic WMPCs Generation}

\begin{figure}
    \centering
    \includegraphics[width=0.75\linewidth]{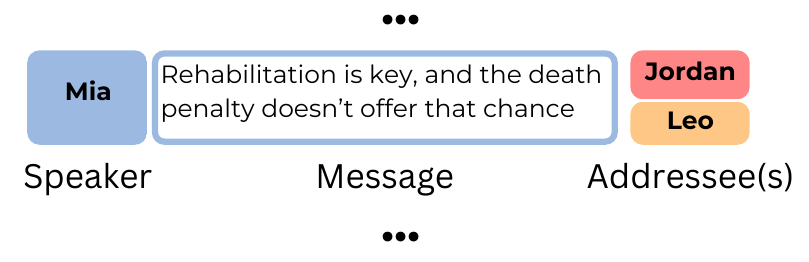}
    \caption{Example of a turn in a synthetic WMPC.}
    \label{fig:example_turn}
\end{figure}

In our framework, a  Written Multi-Party Conversation (WMPC) consists of an ordered sequence of turns, where each turn includes the speaker information (\textit{who} wrote the turn), the message (\textit{what} the textual content of the turn is), and the addressees (to \textit{whom} the turn is directed), see for example Figure \ref{fig:example_turn}. 
In this section, we first introduce the two generation strategies we test (Section \ref{subsec:modalities_pres}), followed by the topics chosen for the WMPCs (Section \ref{subsec:topic_to_discuss}), and finally the constraints specified in the instructions for generating WMPCs (Section \ref{subsec:conv_constr}).

\subsection{Generation Strategies}\label{subsec:modalities_pres}
We test two strategies for generating WMPCs using instruction-based models. 
Our main goal is to determine whether LLMs behave differently when asked to generate a WMPC as a unique narrative compared to acting as an interactive participant within the conversation. With this motivation, we use each LLM in two generation strategies: 

\textbf{One-Long generation strategy (OL).} 
The LLM is prompted to generate the entire conversation in one pass. In this strategy, generation starts with a system input prompt that defines all the constraints and the task, asking then to generate the entire conversation. This strategy follows a one-step, long-generation process, based on a single input context.

\textbf{Turn-by-Turn generation strategy (TT).}  
Here the LLM is prompted to generate the conversation incrementally, provided the conversation history. The model is prompted multiple times to perform one of three tasks:  
\textsc{(i.)} generate a speaker, \textsc{(ii.)} generate interactions between a speaker and addressees (given the candidate speakers/addressees), or \textsc{(iii.)} generate a message (given the interaction).
The process begins with a system prompt specifying the constraints and these three tasks. The model is first prompted to generate each speaker and assign them a stance on a controversial topic. Then, the LLM generates a sequence of interactions and messages (one at a time), iteratively augmenting the WMPC: this means that the context provided to the LLM increases monotonically in size with consecutive turns. 

\subsection{Topics} \label{subsec:topic_to_discuss}

To generate a controlled set of synthetic WMPCs, we identify a set of controversial topics to encourage more polarized and clear statements from speakers based on their assigned stance.
Specifically, following \citet{li-etal-2024-llms-speak}, we select $38$ topics and create two stance statements for each topic: one reflecting a \textit{progressive} perspective and the other a \textit{conservative} perspective. 
Finally, we instruct the LLMs to generate conversations based on each of the resulting 76 statements (see Appendix B at \url{https://arxiv.org/abs/2502.13592}).

\subsection{Conversation Constraints}\label{subsec:conv_constr}

To ensure that the generated conversations feature rich interaction patterns with diverse dynamics, we instruct the model to follow specific constraints, described in the system prompts created for each generation strategy (for details about how this was operationalized in prompts, see Appendix A).

\textbf{Output Format}: to enable automated analysis, the generated output must respect a structured \texttt{JSON} format with all the information needed. So, each generated WMPC must be a dictionary with two main keys, namely  \texttt{conversation} and \texttt{speakers}. The \texttt{conversation} field must include a list of dictionaries, each with specific fields such as  \texttt{speaker}'s name, turn \texttt{message} and \texttt{addressees}, i.e. the list of participants in the conversation to whom the message is directed. The \texttt{speakers} field includes the speaker's \texttt{name} and the \texttt{stance} with respect to the conversation topic. 

\textbf{Interactions}: 
these constraints refer to three requirements in the generated WMPCs -- all speakers appearing in the interactions must be present in the speakers' list (i.e., the LLM should not invent a new speaker half way through the conversation); 
\texttt{addressees} must cover at least once also the role of \texttt{speaker};  self-interactions, i.e. speakers sending a message to themselves, are not admitted.

\textbf{Speaker's Contribution}: all speakers in the \texttt{speakers} field must be authors of at least one turn in the conversation.

\textbf{Number of Speakers}: In order to enable complex interaction structures, each WMPC must involve between 4 and 6 speakers.

\textbf{Number of Messages}: Each generated WMPC must include 15 messages across all speakers, with a maximum of 50 words per message.

\textbf{Speaker's Stance}: We specify the exact number of speakers for each stance (e.g., 2 with the \textit{pro} and 3 with \textit{against} stance). 

We additionally request that the first turn always addresses all participants: this ensures that the generated interaction graph is connected, as required for structural analysis (see Section \ref{subsec:glob}).

\section{Evaluation Framework}\label{sec:analysis}

We design an evaluation framework aimed at assessing different aspects of the generated WMPCs. It is composed of four blocks, which we detail below. 

\subsection{Compliance with Constraints}\label{subsec:filter}

The first dimension considered in the evaluation framework is to what extent the synthetic WMPCs comply with the format and structural constraints given in the prompt. For each generated WMPC, this framework must verify:
\textsc{(i.)} the correctness of the \textit{Output Format}; \textsc{(ii.)} the correctness of the \textit{Interactions}; \textsc{(iii.)} the \textit{Contribution} of each speaker; \textsc{(iv.)} the \textit{Number of Speakers}; \textsc{(v.)} the \textit{Number of Messages}; \textsc{(vi.)} the distribution of the \textit{Stance of the Speakers}.

All the computed values must be compliant with the constraints presented in Section \ref{subsec:conv_constr}. Only for the Number of Messages, we relax the constraint by considering valid WMPCs including less than 15 turns if they contain at least $2$ messages per speaker (on average). Indeed, after a manual check of the generated WMPCs, we noticed that shorter or longer conversations may still represent high-quality data. Each value is computed separately and then used to identify how many WMPCs comply with \textit{all} these constraints.

\begin{figure}
    \centering
    \includegraphics[width=0.78\linewidth]{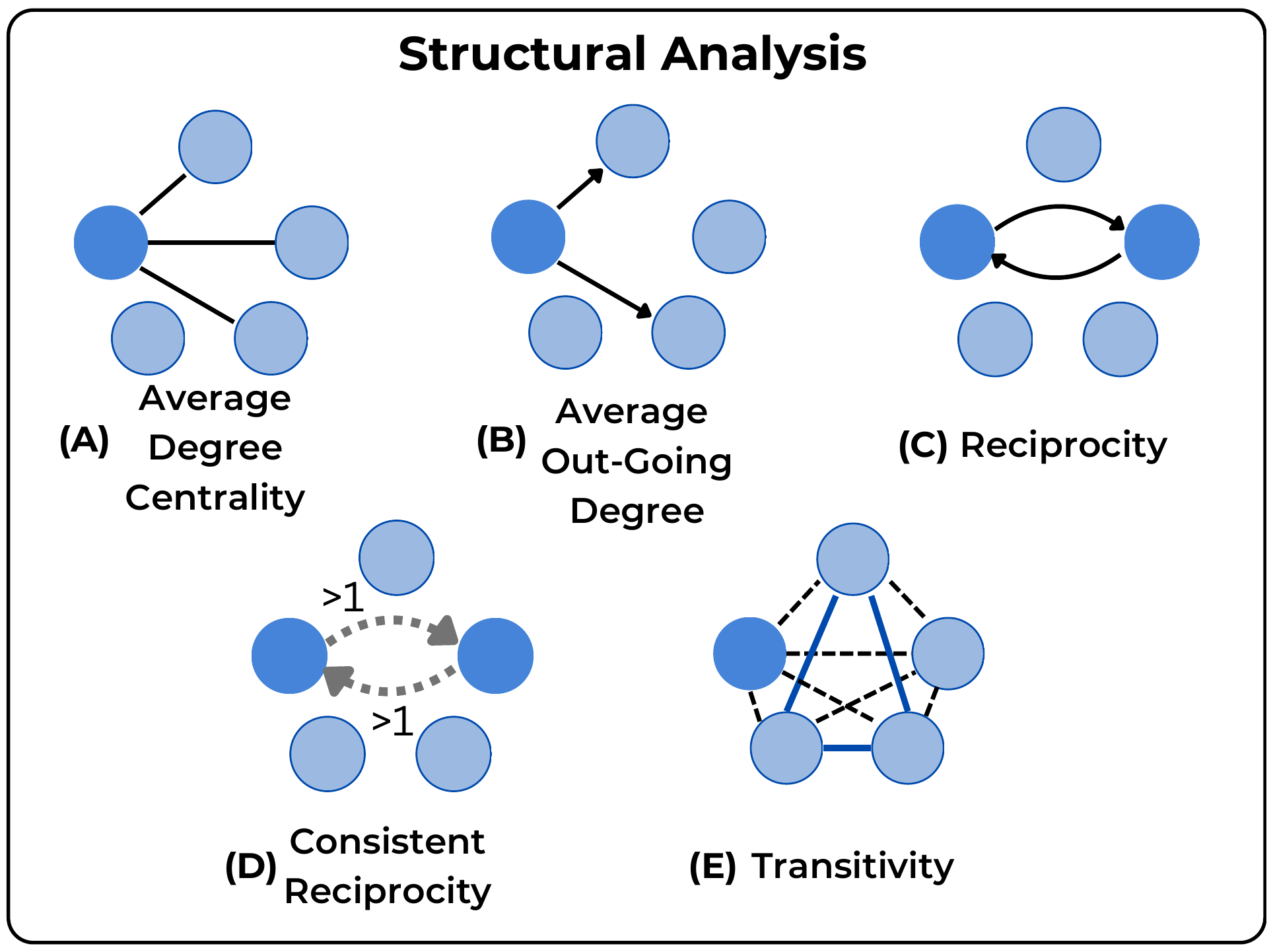}
    \caption{Overview of the metrics considered in our structural analysis.}
    \label{fig:struct_metrics}
\end{figure}

\subsection{Analysis of Language Variability}\label{subsec:content_analysis}

A key risk for synthetic datasets is to suffer from low linguistic variability, due to repetitive examples obtained when using similar prompts (even if stochastic decoding is used), an issue already highlighted for dialogical settings 
\cite{occhipinti-etal-2024-fine}.
On the other hand, while generated WMPCs should ideally be lexically rich, they should also be semantically coherent, i.e. different WMPCs about the same topic should exhibit a certain degree of semantic similarity.  

To control for these aspects, we compute the following three metrics:

\textbf{Repetition Rate} \cite{bertoldi-etal-2013-cache}, which has already been used in synthetic conversational scenarios in the past \citep{bonaldi-etal-2022-human}, measures the rate of non-singleton n-grams within a cluster of WMPCs.  

\textbf{String Similarity} between pairs of turns is computed using thefuzz library\footnote{\url{https://github.com/seatgeek/thefuzz}} and is based on Lehvenstein distance. 

\textbf{Semantic Coherence} between pairs of turns is computed by first embedding each turn with SentenceBERT-all-MiniLM-L6-v2 \cite{reimers-gurevych-2019-sentence} and then calculating pairwise cosine similarity.

We first compute the above metrics at the level of topics (i.e., across all WMPCs generated for the same topic) and then average topic-level scores. 
We provide further details on score computation in Appendix B in the extended version of this paper available on arXiv.

\subsection{Interaction Structure Analysis}\label{subsec:glob}

To describe and quantify the structural complexity of interactions in the generated WMPCs, we compute a series of network metrics, focusing on node-level properties, dyads (pairs of nodes) and triads (triplets of nodes), according to standard practices in interaction network analysis \cite{pauksztat2011speaks, felmlee2021dyads}. 
Following \citet{penzo-etal-2024-llms}, we represent WMPC interactions with an \textit{unweighted undirected graph} $G_{u}$ and a \textit{weighted directed graph} $G_d$, where the weight of an edge corresponds to the number of messages sent in the direction of the edge. 

To measure the average activity of a node in the conversation, we compute two metrics. First, the \textbf{Average Degree Centrality} in $G_u$, denoted as $deg_{avg}(G_{u})$, represents the average number of speakers each participant interacts with, regardless of direction. Second, the \textbf{Average Out-going Degree} in $G_d$, denoted as $outdeg_{avg}(G_{d})$, captures the average number of speakers each participant interacts with having a specific direction. Figure \ref{fig:struct_metrics} provides a visual representation of $deg_{avg}(G_{u})$ (graph A) and $outdeg_{avg}(G_{d})$ (graph B). Both averages are computed across all the nodes in the conversation and normalized according to their maximum possible values.

When two speakers, $s_1$ and $s_2$, reply to each other, they form a \textit{cycle} \cite{coletto2017motif}, represented by a directed edge $e_1$ from $s_1$ to $s_2$ and a reciprocal edge $e_2$ from $s_2$ to $s_1$ (graph C in Figure \ref{fig:struct_metrics}). If this back-and-forth exchange continues multiple times, the edge weights $w(e_1)$ and $w(e_2)$ will both become $>$ $1$. We refer to such recurring exchange as \textit{consistent cycles} (graph D in Figure \ref{fig:struct_metrics}). Based on this, we compute the \textbf{Reciprocity} $R(G_{d})$, i.e., the total number of cycles between two nodes over all pairs of nodes in $G_d$, and the \textbf{Consistent Reciprocity} $R^w(G_{d})$, i.e. the number of consistent cycles between two nodes over all pairs of nodes in $G_d$. Finally, to quantify how often speakers build ``triads'' of interactions, we compute the \textbf{Transitivity} $T(G_{u})$ (graph E in Figure \ref{fig:struct_metrics}), i.e., the number of fully connected subgraphs of size 3 divided by the total number of different subgraphs of the same size in an undirected graph. 

For all these metrics, higher values indicate more complex interactions in a conversation. Indeed, higher reciprocity (consistent or not) suggests more frequent back-and-forth exchanges. Again, higher average degree values means that speakers engage with more participants, while greater transitivity reflects denser connections, leading to the creation of more interconnected speaker groups \cite{pauksztat2011speaks}.

\subsection{Qualitative Evaluation}\label{subsec:hum_eval}

As a final assessment, we evaluate WMPCs qualitatively. We  run both a small-scale human evaluation and an ``LLM as a judge'' assessment \cite{gu2024survey} for a large-scale analysis.  

We ask two expert human annotators and an LLM to rate a given WMPC along the following dimensions (inspired by \citealp{chen-etal-2023-places}) using a Likert Scale from $1$ to $5$: \textsc{(i.)} \textit{naturalness}, i.e., the quality of the overall flow, tone, and word choice in the conversation; 
\textsc{(ii.)} \textit{argumentability}, i.e., how well the conversation presents reasoned and well-argued positions;
\textsc{(iii.)} \textit{speaker's  stance consistency}, i.e., whether all speakers maintain the stance assigned at the beginning of the conversation; 
\textsc{(iv.)} \textit{speaker's  stance evolution}, i.e., whether speakers demonstrate a realistic and logical evolution of their stance during the conversation or maintain their stance consistently;
\textsc{(v.)} \textit{addressee correctness}, i.e., whether the assigned addressees align with the conversation context and are logically appropriate; 
\textsc{(vi.)} \textit{addressee preciseness}, i.e., whether addressees are precise and contextually appropriate (messages should target the smallest relevant group of individuals). For further details see Appendix D and E.

\section{Experimental Settings}

\begin{table*}[ht!]
\centering
\small
{
\begin{tabular}{|l|cc|cc|cc|cc|}
\hline
\textbf{Model}     & \multicolumn{2}{c|}{\textbf{Llama3.1}} & \multicolumn{2}{c|}{\textbf{Qwen2.5}} & \multicolumn{2}{c|}{\textbf{Ministral}} & \multicolumn{2}{c|}{\textbf{OLMo2}}  \\ 
\hline
\textit{Generation strategy}&  \textit{OL} &  \textit{TT} &  \textit{OL}   & \textit{TT} & \textit{OL}  &  \textit{TT} & \textit{OL} & \textit{TT} \\ 
\hline
Output Format     & 78.97  & 97.00  & 90.78  & \textbf{99.58 } & 15.64  & 35.01  & 0.43  & 91.16  \\
Interactions      & 78.91  & 93.49  & 90.72  & \textbf{99.52 } & 15.61  & 13.18  & 0.43  & 70.82  \\
Number of Messages& 78.93  & 70.25  & 90.66  & \textbf{99.57 } & 15.57  & 13.10  & 0.43  & 71.68  \\
Number of Speakers& 29.56  & 97.00  & 39.18  & \textbf{99.57 } & 10.22  & 13.04  & 0.21  & 71.88  \\
Stance of the Speakers & 19.66  & \textbf{96.81 } & 22.95  & 84.03  & 4.42  & 1.04  & 0.09  & 62.11  \\
Contribution      & 72.87  & \textbf{95.29 } & 84.80  & 90.43  & 15.53  & 18.20  & 0.16  & 30.08  \\
\hline
All Constraints & 15.16  & 66.52  & 20.32  & \textbf{77.72 } & 4.34  & 0.87  & 0.04  & 19.39  \\
\hline
\end{tabular}
}
\caption{Number of generated WMPCs that are compliant with each constraint (percentage on the full set of $102\,600$ generations) for each LLM and strategy (i.e. OL = One-Long generation, TT = Turn-by-Turn generation). The final percentage of WMPCs (last row) is the percentage of generations that satisfy all constraints.}
\label{tab:constraints_satisf_percent}
\end{table*}

To generate synthetic WMPCs, we compare four different instruction-based models, chosen for their comparable parameter sizes and compatibility with the same prompt design. The models include  Llama3.1-8B-Instruct \cite{dubey2024llama3herdmodels}, Qwen2.5-7B-Instruct \cite{yang2024qwen2technicalreport}, Ministral-8B-Instruct\footnote{\url{https://mistral.ai/news/ministraux/}}, and OLMo-2-7B-Instruct \cite{olmo20242olmo2furious}.
For each generation strategy (One-Long or Turn-by-Turn, see Section \ref{subsec:modalities_pres}) we develop three distinct system prompts combining a more or less schematic task description and different examples of the output format. For details we refer to Appendix A. For each combination of constraints, topic and system prompt, we generate $75$ conversations to account for the potential variety of structures.
In total we obtained $102\,600$ synthetic WMPCs for each model and generation strategy.

\section{Evaluation Results}

We evaluate the generated WMPCs for each dimension of the evaluation framework (Section  \ref{sec:analysis}).

\subsection{Evaluation of Compliance with Constraints}\label{subsec:result_constr}

We first address \textbf{RQ(1)}, aimed at assessing whether synthetic WMPCs can comply with the predefined constraints described in Section \ref{subsec:filter}. The results of the analysis are reported in Table \ref{tab:constraints_satisf_percent}. We compare the output generated by the four different LLMs, each following two strategies for generation (i.e. OL vs. TT). We report the percentage of generated WMPCs, out of the $102\,600$ in the initial set, that were generated in compliance with the given constraint.

This first evaluation shows that Qwen2.5 is the best model to comply with the constraints, followed by Llama3.1. Indeed, focusing on the best generation strategy, $77.72\%$ of the WMPCs generated by the former comply with all constraints, while for Llama 3.1 this percentage drops to $66.52\%$. Ministral and OLMo2, instead, fail to satisfy all constraints in the vast majority of generated conversations. Concerning the generation strategy, TT generation is overall better at complying with almost all the constraints.

The constraints where most settings encountered significant challenges were the \textit{Number of Speakers} and  \textit{Stance of the Speakers}. However, TT seems to be able to mitigate these issues for LLMs except for Ministral.
Based on these findings, in the remainder of this work we will focus on Llama3.1 and Qwen2.5 and perform all analyses on the subset of WMPCs that satisfy all constraints.

\subsection{Results of Language Variability}

\begin{table}[t]
    \centering
    \small
    \begin{tabular}{|l|cc|cc|}
    \hline
  \textbf{Model} & \multicolumn{2}{c|}{\textbf{Llama3.1}} & \multicolumn{2}{c|}{\textbf{Qwen2.5}}  \\ \hline
  \textit{Gener. Strategy}  & \textit{OL}      & \textit{TT}       & \textit{OL}    & \textit{TT} \\ \hline
          Avg. \# words & $11.94$ & $26.58$ & $9.67$ & $14.15$\\\hline

        RepetitionRate ($\downarrow$) & $18.08$ & $11.07$ & $14.43$ & $13.35$\\
        StringSimilarity ($\downarrow$) & $65.51$ & $53.88$ & $63.22$ & $58.38$\\
         SemanticCoher. ($\uparrow$) & $0.606$ & $0.636$ & $0.588$ & $0.604$\\

        \hline
    \end{tabular}
    \caption{Results of language variability analysis.}
    \label{tab:content_analysis}
\end{table}

Table \ref{tab:content_analysis} summarizes the results on language variability (as described in Section \ref{subsec:content_analysis}). 
The analysis shows that linguistic variability at surface level is lower when WMPCs are generated in a single pass (OL generation) and also semantic coherence is lower compared to TT generation for both models, i.e., Llama3.1 and Qwen2.5. This is probably due to the fact that in TT settings, the LLM is explicitly required to generate a turn by taking into account what immediately precedes it, building a coherent conversation step by step. 
Llama3.1 generates less repetitive WMPCs at surface level, despite their turns being on average longer than Qwen2.5's. Also semantic coherence is generally better for Llama3.1. 

\begin{figure*}
    \centering
    \includegraphics[width=0.79\linewidth]{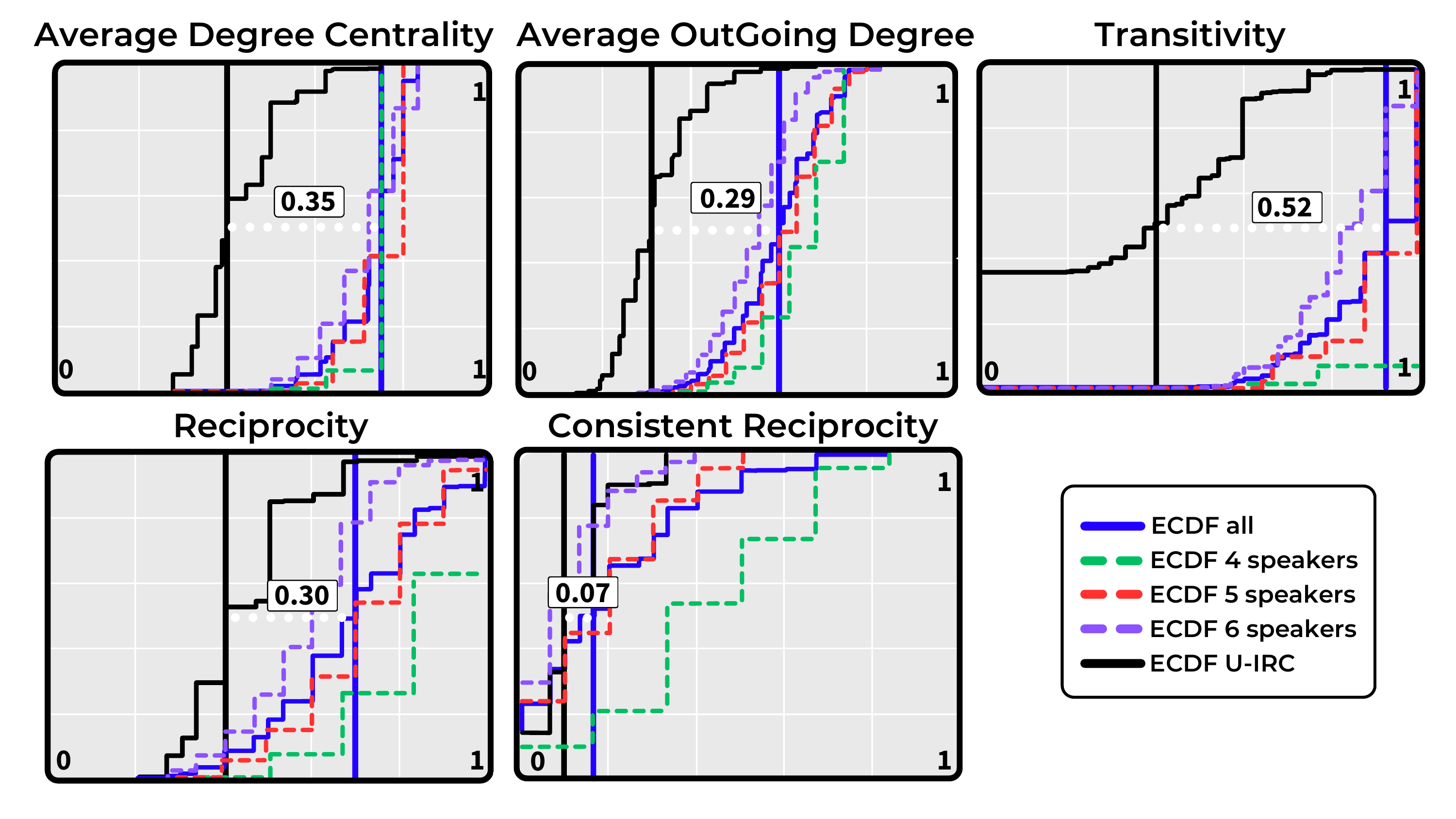}
    \caption{Empirical Cumulative Density Function (ECDF) of structural analysis of synthetic WMPCs from Qwen2.5-TT. All the boxes go from $0$ to $1$ on both vertical axis (density) and horizontal axis (value of the metric).}
    \label{fig:glob_struct}
\end{figure*}

\subsection{Results of Structure Analysis}\label{subsec:global-analysis-expl}
 We report the results of the structure analysis for Qwen2.5 -- TT, i.e. the model providing the highest number of synthetic WMPCs, in Figure \ref{fig:glob_struct}. Results for Qwen2.5 -- OL and for Llama3.1 exhibit similar patterns, which are detailed in the Appendix B.

Since one of our goals is to assess how synthetic WMPCs compare to \textit{real} WMPCs in terms of structural complexity, we perform the same structure analysis on $13\,714$ WMPCs extracted from the UbuntuIRC dataset \citep{ouchi-tsuboi-2016-addressee}, a widely used corpus of conversations from an online forum about software issues and troubleshooting. This subset was extracted using the strategy in \citet{penzo-etal-2024-llms} to obtain all non-overlapping conversations with 15 messages and 4, 5, or 6 speakers, ensuring each conversation formed a single connected-component (in terms of interaction graph). 
For each of the five network metrics introduced in Section \ref{subsec:glob}, we plot in Figure \ref{fig:glob_struct} the Empirical Cumulative Density Function (ECDF) obtained by analysing synthetic WMPCs with 4, 5 or 6 speakers (i.e. nodes) and on all generated WMPCs, and we compare them with ECDF for UbuntuIRC. 

For all metrics, higher values indicate more complex interactions. As shown by the median values, the UbuntuIRC dataset consistently exhibits lower values across all statistics. Compared to UbuntuIRC, speakers in our synthetic WMPCs tend to interact with more participants. Also, pairs of speakers tend to have more back-and-forth dynamics and groups of speakers tend to be more interconnected.
Additionally, in our dataset, the distribution of conversations with varying numbers of participants closely mirrors the overall average, with no notable deviations. This finding holds for all the model-strategy combinations and all metrics.

\subsection{Qualitative Evaluation}\label{subsec:qualitative_res}
The last analysis focuses on the quality of the generated conversations and is conducted both manually and automatically. 
Ideally, using LLM-as-a-judge would allow us to quickly evaluate all synthetic WMPCs with limited effort. However, we need to assess the quality of this automatic multi-dimensional evaluation. 
So, we first select $96$ WMPCs ($24$ per model and generation strategy) via stratified sampling balanced across topic and stance. 

We then ask two  human annotators with extensive experience in  linguistic annotation
to evaluate for each WMPC the six dimensions described in Section \ref{subsec:hum_eval} (addressee correctness, stance consistency, etc.).

The average values assigned to each dimension on a Likert scale between $1$ (poor quality) and $5$ (perfect quality) on the 96 WMPCs are reported in Table \ref{tab:human_res}. We observe that all dimensions have been evaluated positively, especially \textit{Naturalness} and \textit{Speaker's Stance Evolution}. The most challenging dimension is \textit{Addressee Preciseness}, which is the only dimension with an average score below $4$ for all combinations. Neither of the two LLMs is consistently better and neither of the generation strategies (OL vs. TT) is superior to the other w.r.t all evaluation dimensions. The inter-annotator  agreement, measured via Krippendorff’s alpha \cite{krippendorff1computing} and Spearman's correlation on all 96 WMPCs, shows high agreement on the stance-based dimension, medium for addressee-based ones and lower for the content-based dimensions. We provide more details in Appendix D.

\begin{table}
    \centering
    \small
    {
    \begin{tabular}{|l|cc|cc|}
    \hline
  \textbf{Model} & \multicolumn{2}{c|}{\textbf{Llama3.1}} & \multicolumn{2}{c|}{\textbf{Qwen2.5}}  \\ \hline
  \textit{Generation Strategy} & \textit{OL}      & \textit{TT}       & \textit{OL}    & \textit{TT}  \\ \hline
        Naturalness  & $\textbf{4.46}$ & $4.29$ & $4.33$ & $4.00$ \\
        Argumentability  & $3.98$ & $\textbf{4.17}$ & $3.83$ & $3.52$ \\
        Addressee Correctness  & $4.02$ & $4.10$ & $3.92$ & $\textbf{4.21}$ \\
        Addressee Preciseness  & $3.65$ & $\textbf{3.94}$ & $3.81$ & $3.52$ \\
        Stance Consistency  & $4.04$ & $3.65$ & $3.60$ & $\textbf{4.21}$ \\
        Stance Evolution& $4.29$ & $\textbf{4.73}$ & $4.33$ & $4.42$ \\ \hline
    \end{tabular}
     }
    \caption{Average results between the two human annotators on 96 WMPCs (24 for each model-strategy combination).}
    \label{tab:human_res}
\end{table}

We complement this manual evaluation with a large-scale automatic LLM-as-a-judge evaluation with OpenAI's o3-mini model.\footnote{\url{https://openai.com/index/openai-o3-mini/}} We first assess whether it can be reliably used to evaluate all six dimensions above. We therefore launch LLM-as-a-judge on the same 96 WMPCs which were manually evaluated and measure human-LLM agreement (full details in Table 10 in Appendix D).

While Spearman's correlation highlights a positive correlation between LLM and both human annotators on all dimensions except for \textit{Addressee Preciseness}, Krippendorf's alpha results are less consistent. Only the \textit{Speaker Stance Consistency}, i.e. whether the speakers comply with the assigned stance when entering the conversation, shows an extremely high agreement and correlation (Krippendorf's alpha $0.80$, Spearman's correlation $0.76$/$0.78$).

We therefore carry out a large-scale evaluation only on the stance-based dimensions\footnote{We use  LLM-as-a-judge also on the \textit{Speaker's Stance Evolution}, where correlation was still highly statistical significant.} using LLM-as-a-judge on 800 conversations (200 per model and generation strategy). Results are reported in Table \ref{tab:llm_judge_res} and, similar to the human evaluation, show that Llama3.1 and Qwen2.5 are comparable in terms of performance and that they are able to generate WMPCs that present realistic evolution of speakers' stance with both generation strategies.

 \begin{table}[ht]
    \centering
    \small
    \begin{tabular}{|l|cc|cc|}
    \hline
  \textbf{Model} & \multicolumn{2}{c|}{\textbf{Llama3.1}} & \multicolumn{2}{c|}{\textbf{Qwen2.5}} \\ \hline
    \textit{Generation Strategy}  & \textit{OL}      & \textit{TT}       & \textit{OL}    & \textit{TT}  \\ \hline
        Stance Consistency  & $\textbf{4.15}$ & $3.76$ & $3.99$ & $3.64$ \\
        Stance Evolution  & $4.64$  & $4.46$ & $4.62$ & $\textbf{4.68}$\\ \hline
    \end{tabular}
    \caption{Results with LLM as a judge on 800 WMPCs (200 for each model-strategy combination).}
    \label{tab:llm_judge_res}
\end{table}

\section{Discussion}

The analyses from the previous sections allow us to address the three research questions from Section \ref{sec:intro}. 
With respect to \textbf{RQ(1)}, targeting the possibility to generate synthetic WMPCs following predefined constraints, our evaluation shows that models with comparable parameter sizes can yield very different performances. In this respect, Qwen2.5 is by far the best performing LLM followed by Llama3.1. Indeed, it is able to generate $77.72\%$ of WMPCs compliant with \textit{all} the constraints provided in the prompt. 
The reason behind this difference in performance cannot be clearly identified but it likely depends on the quality of pretraining data. 
Looking at other dimensions, however, there is no clear winner between Qwen2.5 and Llama3.1. Although Llama3.1 generates less repetitive and semantically more coherent WMPCs, our qualitative evaluation does not favor either model. 

As regards \textbf{RQ(2)}, aimed at finding the best generation strategy between OL and TT, we observe that generating WMPCs in a Turn-by-Turn fashion is consistently better in terms of compliance with given constraints. This can be related to recent advancements in handling long contexts: generating shorter, multi-step outputs can be more precise and reduce errors compared to relying on a single, long-generation output. However, this advantage comes at the cost of longer computational times (in our experiments, TT took from $4$ to $8$  times more than OL, see Appendix C). Using TT reduces also the repetitiveness of WMPCs while generating conversations that are more semantically coherent than OL. Our qualitative evaluation, in contrast, renders TT and OL similarly viable. 

To address \textbf{RQ(3)}, concerning how we can effectively evaluate the quality of generated WMPCs along different dimensions, we present a framework composed by four evaluation blocks, each targeting a specific aspect of WMPCs. Beside linguistic variety, coherence and qualitative dimensions such as naturalness and stance evolution, we introduce a novel assessment of the structure of synthetic WMPCs. We consider five network metrics and compute empirical cumulative density function to compare them with the same values calculated from real WMPCs. We show that it is possible to steer the interaction structure in generated conversations, which paves the way to the large-scale creation of high-quality WMPCs with much more complex interactions than what social media datasets offer.

\section{Limitations and Ethical Considerations}

Our work presents some limitations. First of all, we focus only on English, and the topics we select are typical of US-centric  polarized debates such as universal healthcare, right to abortion and death penalty. It is possible that precisely because of these divisive topics, speakers in generated WMPCs were able to discuss in a consistent way with respect to the assigned stance. In the future, it would be interesting to extend our analysis also to topics on which speakers can have more nuanced views, that are probably more challenging for LLMs to imitate. Moreover, we generated WMPCs with 4, 5 or 6 speakers, and with a length of 15 turns. It may be worth investigating whether looser constraints, allowing more or less speakers, or longer and shorter conversations, can lead to the creation of more ``natural'' WMPCs and whether the evaluation results would still hold.

One of the main reasons behind research on synthetic WMPCs is the need to comply with privacy concerns, especially when working with conversations extracted from social media, to alleviate ethical issues related to sharing personal information online. Still, we acknowledge that the problem is not fully solved since basically all best performing LLMs are currently trained on social media data, and synthetic WMPCs could include personal data as well \cite{li2024llm}. Also, the creation of synthetic WMPCs is not exempt from possible negative impact, for instance when used for training malicious agents in social conversation scenarios. 

Finally, the models we use are openly available and accessible to anyone. Our approach does not involve any form of forced jailbreak or manipulation to elicit toxic behavior. While it is challenging to verify the complete absence of toxic language across large synthetic datasets, in the manual evaluation of the dialogues we did not observe such phenomena.

\section{Conclusion}
The creation of WMPCs widely relies on  social media data because of its abundance and accessibility. Due to platform constraints and inherently asynchronous communication, however, such datasets poorly reflect the structural diversity of natural WMPCs.

In this work, we investigated the viability of generating varied MPCs with LLMs, showing that (some) LLMs can indeed generate WMPCs that conform to structural constraints (e.g., number of speakers and their stances). Models such as Llama3.1 and Qwen2.5 can yield high-quality WMPCs under varied constraints, both when prompted to \textsc{(i.)} generate the whole WMPC at once or \textsc{(ii.)} one turn at a time, given all preceding turns in context. 
 
This makes LLMs suitable for synthesizing large-scale datasets for various types of conversations, addressing the diversity of real-world WMPCs. Synthesized data can then be further leveraged to fine-tune smaller models for various discriminative tasks (e.g., next speaker or addressee prediction). Our future efforts will exactly focus on synthesizing use-case-specific WMPCs and evaluating their utility when used as fine-tuning data for smaller discriminative models.

\section*{Acknowledgments}
We thank Sebastiano Vecellio Salto for his contribution to the evaluation activities. The work of BL was partially supported by the NextGenerationEU Horizon Europe Programme, grant number 101120237 - ELIAS and grant number 101120763 - TANGO. BL and ST were also supported by the PNRR project FAIR - Future AI Research (PE00000013). NP’s activities are part of the network of excellence of the European Laboratory for Learning and Intelligent Systems (ELLIS).
The work is a result of his research visit at the Chair for Natural Language Processing, Center For Artificial Intelligence and Data Science, University of W\"urzburg, Germany, under the supervision of GG. The visiting has been partially funded by the Erasmus+ Traineeship programme. The work of GG was partially supported by the Alcatel-Lucent Stiftung and Deutsches Stiftungszentrum through the grant ``Equitably Fair and Trustworthy
Language Technology'' (EQUIFAIR, Grant Nr. T0067/43110/23).  

\bibliography{aaai2026}


\setcounter{secnumdepth}{1}

\newpage

\appendix

\section{System Prompts}\label{appendix:system_prompts}

\begin{figure*}
    \centering
    \includegraphics[width=0.7\linewidth]{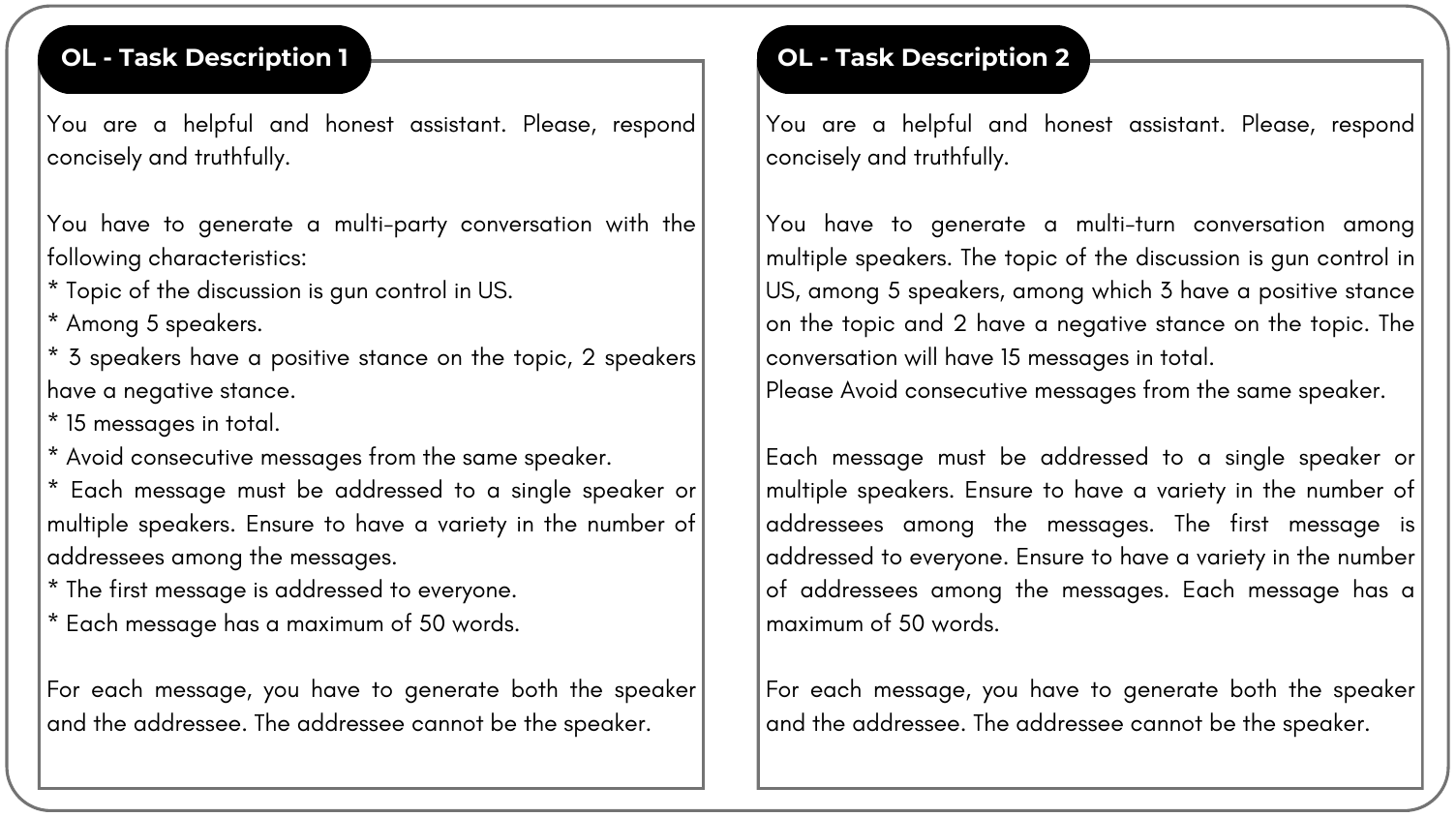}
    \caption{The two versions of \texttt{Task Description} for the One-Long generation strategy}
    \label{fig:ol_system}
\end{figure*}

\begin{figure*}
    \centering
    \includegraphics[width=0.65\linewidth]{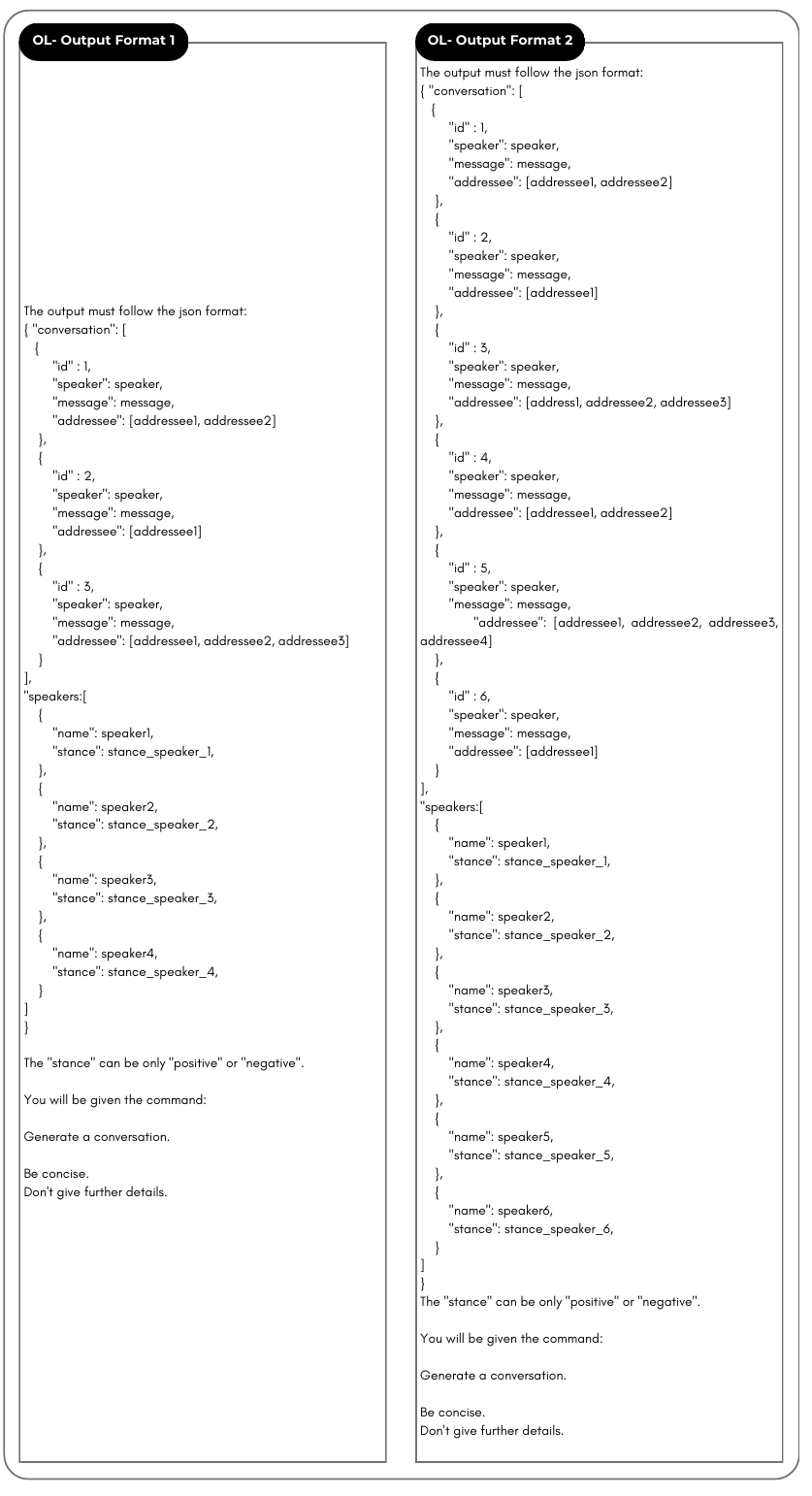}
    \caption{The two examples of \texttt{Output Format} for the One-Long generation strategy}
    \label{fig:ol_output}
\end{figure*}

\begin{figure*}
    \centering
    \includegraphics[width=0.7\linewidth]{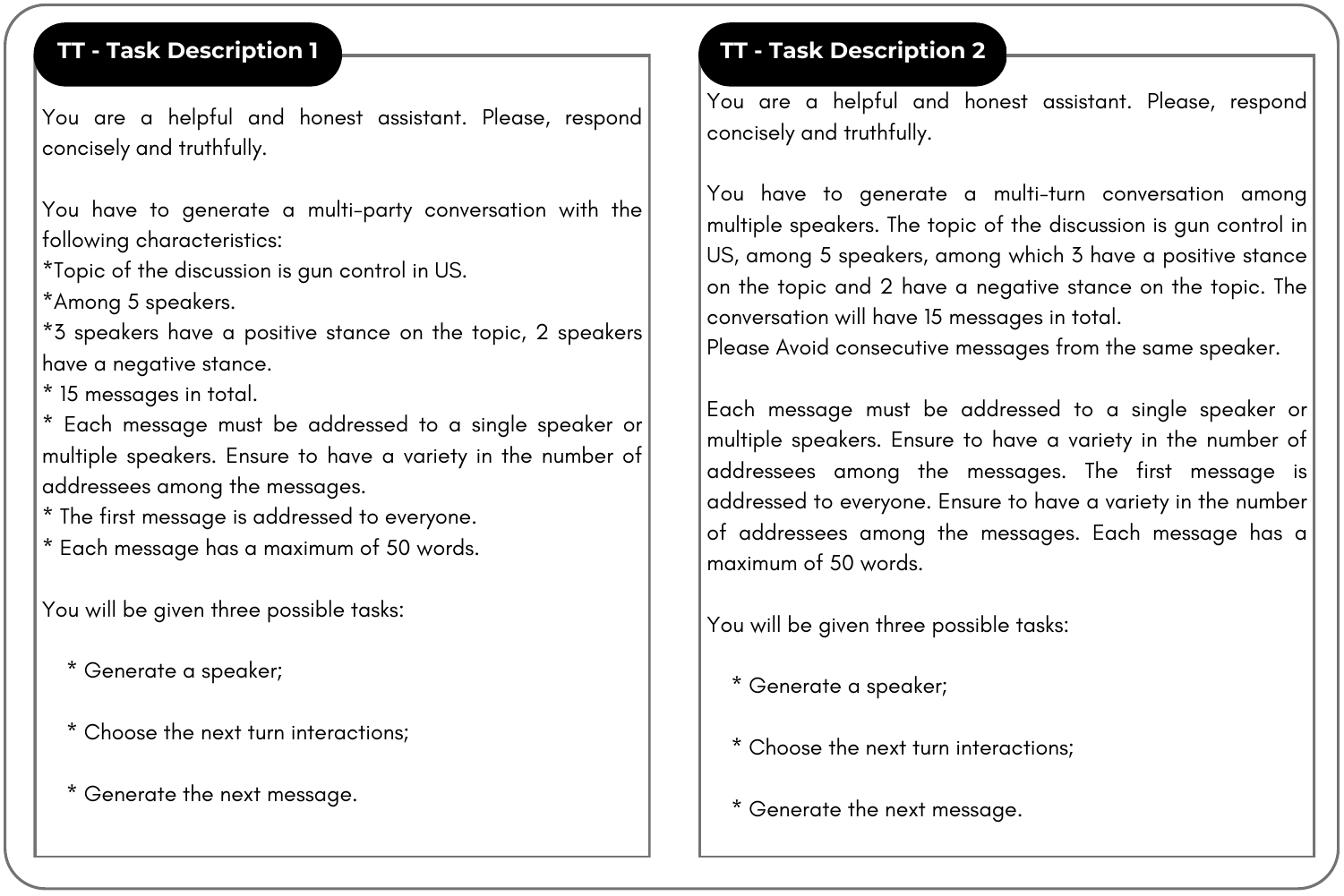}
    \caption{The two versions of \texttt{Task Description} for the Turn-by-Turn generation strategy}
    \label{fig:tt_system}
\end{figure*}

\begin{figure*}
    \centering
    \includegraphics[width=0.7\linewidth]{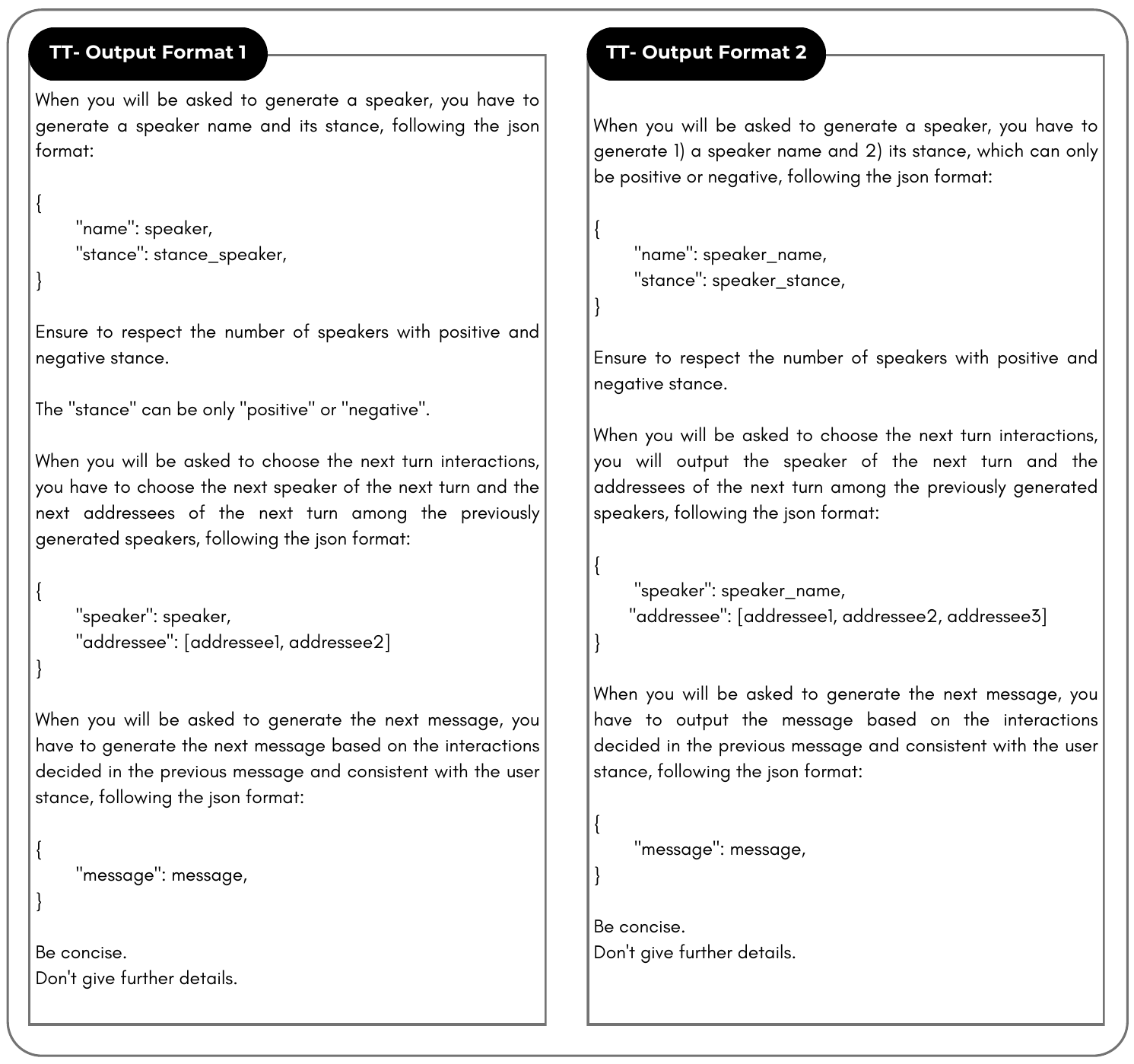}
    \caption{The two examples of \texttt{Output Format} for the Turn-by-Turn generation strategy}
    \label{fig:tt_output}
\end{figure*}

We report in Figure \ref{fig:ol_system} and Figure \ref{fig:ol_output} respectively the two versions of the \texttt{Task Description} and the two versions of the \texttt{Output Format} (same structure, just different example) for the One-Long (OL) generation strategy. In Figure \ref{fig:tt_system} and Figure \ref{fig:tt_output} we report the same for the Turn-by-Turn (TT) generation strategy. 

For both strategies, the three System Prompts are obtained by concatenating: \textsc{(i.)} \texttt{Task Description 1} + \texttt{Output Format 1}; \textsc{(ii.)} \texttt{Task Description 2} + \texttt{Output Format 1};  \textsc{(iii.)} \texttt{Task Description 1} + \texttt{Output Format 2}. 

\begin{figure*}
    \centering
    \includegraphics[width=0.9\linewidth]{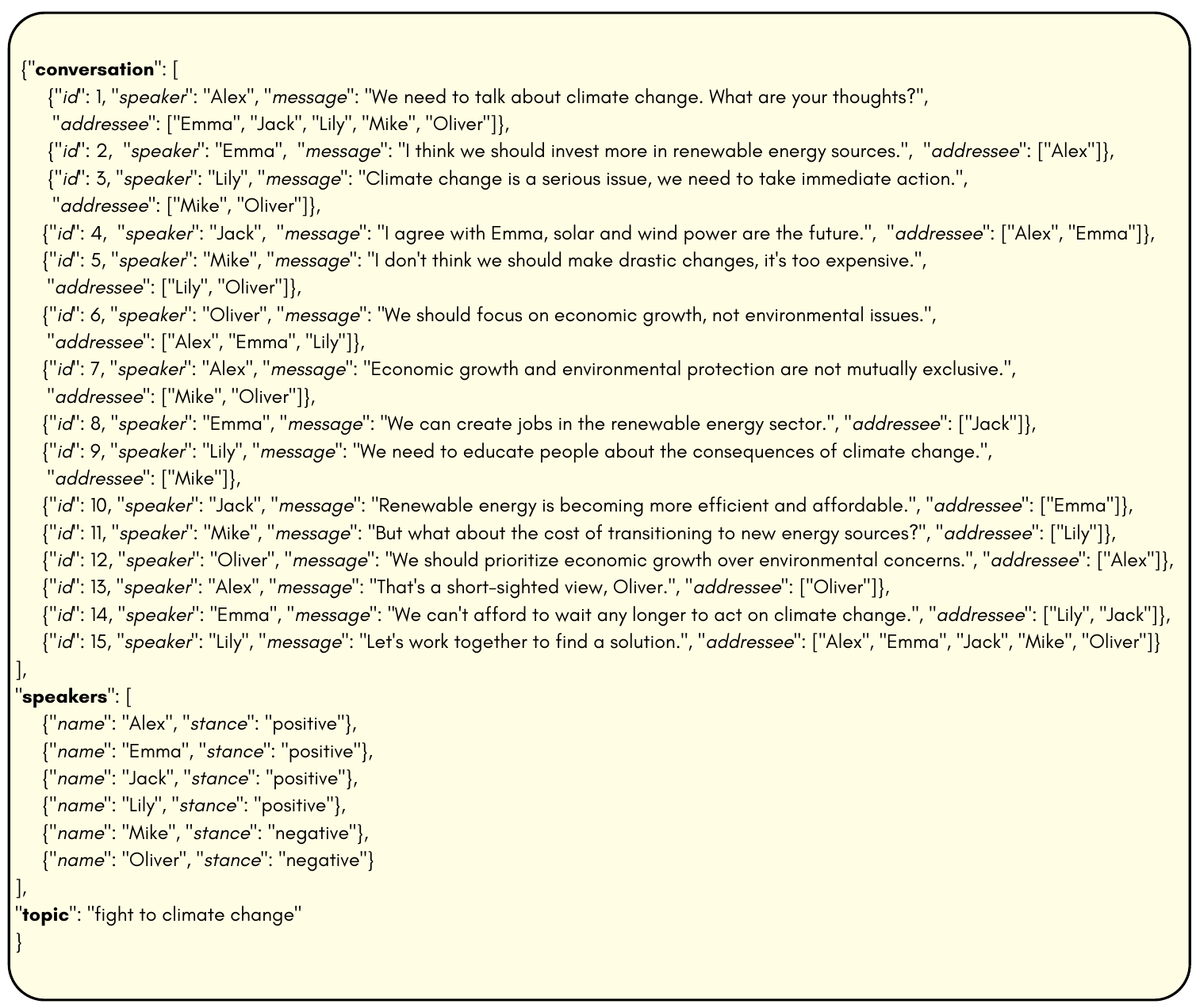}
    \caption{Example of Written Synthetic Multi-Party Conversation}
    \label{fig:conv_example}
\end{figure*}

Finally, in Figure \ref{fig:conv_example} we report a synthetic Written Multi-Party Conversation according to the guidelines (the \texttt{topic} field has been added in post-processing for context). In the System Prompt, the  assignment of stance among the speakers has $6$ possible  distributions (pro-topic vs counter-topic): $2$ vs $2$, $3$ vs $2$, $2$ vs $3$, $2$ vs $4$, $3$ vs $3$, $4$ vs $2$.

\section{Further Analysis}\label{appendix:further_analysis}

\subsection{List of Topics} \label{app:topic_to_discuss}

The topics on which LLMs were prompted to generate WMPCs (see Section \ref{subsec:topic_to_discuss}) are reported in Table \ref{tab:progressist_conservatist}. They consist in $76$ paired statements based on $38$ topics, in which each statement reflects a more conservative or progressive point of view with respect to the given topic. The statements are created by avoiding potential biases such as framing statements negatively or using specific terms exclusively in one category.

The $38$ topics were manually selected from the ones provided by \citet{li-etal-2024-llms-speak} (freely available in the related Github repository).\footnote{\url{https://github.com/tianyi-lab/DEBATunE}}
Specifically, we picked the most polarizing topics, in order to foster more clear-cut stances during the generation of WMPCs. 

\begin{table*}[ht]
\centering
\small
\begin{tabular}{|p{7cm}|p{7cm}|}
\hline
\textbf{Progressive} & \textbf{Conservative} \\ \hline
ban of targeted killing & allowance of targeted killing \\ \hline
ban of the death penalty & allowance of the death penalty \\ \hline
recognition of the right to abortion & ban of abortion \\ \hline
recognition of the right to euthanasia & ban of euthanasia \\ \hline
recognition of Palestinian state & non-recognition of Palestinian state \\ \hline
ban of mandatory military service & mandatory military service \\ \hline
ban of nuclear weapons & support for nuclear weapons \\ \hline
mandatory sex education in schools & optional sex education in schools \\ \hline
guarantee of online teaching & mandatory in-person teaching \\ \hline
fight to climate change & opposition to regulations for action on climate change \\ \hline
incentives for renewable energy & incentives for energy from fossil fuels \\ \hline
ban of facial recognition technology & incentives for facial recognition technology \\ \hline
incentives for AI research & opposition to AI research incentives \\ \hline
mandatory vaccination for children & optional vaccination for children \\ \hline
ban of animal testing & allowance of animal testing \\ \hline
incentives for organ donation & opposition to organ donation incentives \\ \hline
ban of racial profiling & allowance of racial profiling \\ \hline
incentives for immigration and asylum & support to immigration contrast and stricter asylum rules \\ \hline
universal healthcare & support to private healthcare \\ \hline
legalization of marijuana & ban of marijuana \\ \hline
legalization of same-sex marriage & ban of same-sex marriage \\ \hline
legalization of surrogate motherhood & ban of surrogate motherhood \\ \hline
programme for the reduction of the gender pay gap & increase of the gender pay gap in favor of men \\ \hline
limitation to gun ownership & right to unrestricted gun ownership \\ \hline
holocaust remembrance mandatory in schools & optional holocaust remembrance in schools \\ \hline
ban of zoos & support for zoos \\ \hline
protection of endangered species & opposition to endangered species protection \\ \hline
organization of pride parades & ban of pride parades \\ \hline
allowance of tattoos & ban of tattoos \\ \hline
cohabitation of couples before marriage & mandatory marriage before cohabitation \\ \hline
ban of arranged marriages & right to arranged marriages \\ \hline
US staying in NATO & US leaving NATO \\ \hline
Germany staying in EU & Germany leaving the EU \\ \hline
mandatory acceptance of mobile payments & ban of mobile payments \\ \hline
lowering university tuition fees & increase in university tuition fees \\ \hline
mandatory cameras on police officers & freedom of police officers to refuse cameras \\ \hline
freedom of blasphemy & punishment for blasphemy \\ \hline
legalization of adoption by same-sex couples & ban of adoption by same-sex couples \\ \hline
\end{tabular}
\caption{List of topics, paired according to their Progressive and Conservative version.}
\label{tab:progressist_conservatist}
\end{table*}

\subsection{Effects of prompt formulation}
Table \ref{tab:constraints_satisf_percent_llama_qwen} and Table \ref{tab:constraints_satisf_percent_ministral_olmo} report the percentage of WMPCs that are compliant with each constraint, as in Section \ref{subsec:result_constr}, computed over the full set of $34\,000$ WMPCs generated for each combination of System Prompt (as presented in Section \ref{appendix:system_prompts}), model, and generation strategy. The constraints definitions follow those introduced in Section \ref{subsec:conv_constr}, and the results are broken down by both the generation strategies and the three system prompts described in Appendix \ref{appendix:system_prompts}.

As the tables show, there is no universally best-performing prompt across models or strategies. Importantly, restricting evaluation to only the best-performing prompt does not affect the model-strategy combinations' ranking discussed in Section \ref{subsec:result_constr}. However, doing so would ignore the variability introduced by different prompt formulations. For this reason, we report results across all three system prompts in the main body of the paper.

\begin{table*}[ht!]
\centering
\small
{
\begin{tabular}{|l|ccc|ccc|ccc|ccc|}
\hline
\textbf{Model} & \multicolumn{6}{c|}{\textbf{Llama3.1}} & \multicolumn{6}{c|}{\textbf{Qwen2.5}} \\
\hline
\textit{Generation strategy} & \multicolumn{3}{c|}{\textit{OL}} & \multicolumn{3}{c|}{\textit{TT}} & \multicolumn{3}{c|}{\textit{OL}} & \multicolumn{3}{c|}{\textit{TT}} \\
\hline
\textsc{System Prompt} & \textsc{i} & \textsc{ii}  & \textsc{iii}  & \textsc{i}  & \textsc{ii}  & \textsc{iii}  & \textsc{i}  & \textsc{ii}  & \textsc{iii}  & \textsc{i}  & \textsc{ii}  & \textsc{iii}  \\
\hline
Output Format & 76.6  & 74.1  & 86.1  & 97.1  & 97.3  & 96.6  & 85.3  & 92.0  & 95.1  & 99.4  & 99.7  & 99.6  \\
Interactions & 76.6  & 74.1  & 86.1  & 93.5  & 94.5  & 92.4  & 85.2  & 91.9  & 95.0  & 99.3  & 99.7  & 99.6  \\
Number of Messages & 76.6  & 74.1  & 86.1  & 60.9  & 76.9  & 72.9  & 85.2  & 91.9  & 94.9  & 99.4  & 99.7  & 99.6  \\
Number of Speaker & 33.6  & 34.1  & 21.0  & 97.1  & 97.3  & 96.6  & 32.3  & 25.8  & 59.5  & 99.4  & 99.7  & 99.6  \\
Stance of the Speakers & 21.9  & 22.4  & 14.7  & 96.9  & 97.1  & 96.4  & 18.0  & 16.0  & 34.9  & 86.4  & 81.3  & 84.4  \\
Contribution & 73.0  & 67.4  & 78.2  & 94.8  & 96.3  & 94.7  & 79.9  & 88.5  & 86.0  & 89.8  & 91.3  & 90.2  \\
\hline
All Constraints & 18.9  & 18.0  & 8.6  & 57.1  & 73.8  & 68.6  & 15.5  & 14.1  & 31.5  & 79.9  & 75.4  & 77.9  \\
\hline
\end{tabular}
}
\caption{Number of generated WMPCs that are compliant with each constraint for each of the three prompt versions (percentage out of the full set $34\,200$ generations) for Llama3.1 and Qwen2.5, in both strategies (i.e. OL = One-Long generation, TT = Turn-by-Turn generation).}
\label{tab:constraints_satisf_percent_llama_qwen}
\end{table*}

\begin{table*}[ht!]
\centering
\small
{
\begin{tabular}{|l|ccc|ccc|ccc|ccc|}
\hline
\textbf{Model} & \multicolumn{6}{c|}{\textbf{Ministral}} & \multicolumn{6}{c|}{\textbf{OLMo2}} \\
\hline
\textit{Generation strategy} & \multicolumn{3}{c|}{\textit{OL}} & \multicolumn{3}{c|}{\textit{TT}} & \multicolumn{3}{c|}{\textit{OL}} & \multicolumn{3}{c|}{\textit{TT}} \\
\hline
\textsc{System Prompt} & \textsc{i} & \textsc{ii}  & \textsc{iii}  & \textsc{i}  & \textsc{ii}  & \textsc{iii}  & \textsc{i}  & \textsc{ii}  & \textsc{iii}  & \textsc{i}  & \textsc{ii}  & \textsc{iii}  \\
\hline
Output Format & 16.5 & 13.5  & 16.9  & 36.6  & 37.7  & 30.8  & 0.0  & 0.3  & 1.0  & 92.3  & 93.3  & 87.9  \\
Interactions & 16.5 & 13.5  & 16.8  & 13.7  & 12.2  & 13.6  & 0.0  & 0.3  & 1.0  & 83.5  & 67.9  & 61.1  \\
Number of Messages & 16.5 & 13.5  & 16.8  & 13.7  & 12.1  & 13.5  & 0.0  & 0.3  & 1.0  & 84.3  & 69.4  & 61.4  \\
Number of Speakers & 14.1 & 9.8  & 6.9  & 13.6  & 12.0  & 13.5  & 0.0  & 0.1  & 0.6  & 84.0  & 70.3  & 61.3  \\
Stance of the Speakers & 6.1 & 4.4  & 2.8  & 0.9  & 1.2  & 1.1  & 0.0  & 0.1  & 0.2  & 75.8  & 60.8  & 49.7  \\
Contribution & 16.5 & 13.4 & 16.7  & 21.3  & 16.6  & 16.8  & 0.0  & 0.3  & 0.2  & 27.9  & 28.6  & 33.8  \\
\hline
All Constraints & 6.1  & 4.3  & 2.7  & 0.7  & 1.0  & 0.9  & 0.0  & 0.1  & 0.1  & 22.8  & 17.8  & 17.6  \\
\hline
\end{tabular}
}
\caption{Number of generated WMPCs that are compliant with each constraint for each of the three prompt versions (percentage out of the full set $34\,200$ generations) for Ministral and OLMo2, in both strategy (i.e. OL = One-Long generation, TT = Turn-by-Turn generation).}
\label{tab:constraints_satisf_percent_ministral_olmo}
\end{table*}

\subsection{General Statistics of Final Set of Synthetic WMPCs}
\begin{figure*}
    \centering
    \includegraphics[width=0.95\linewidth]{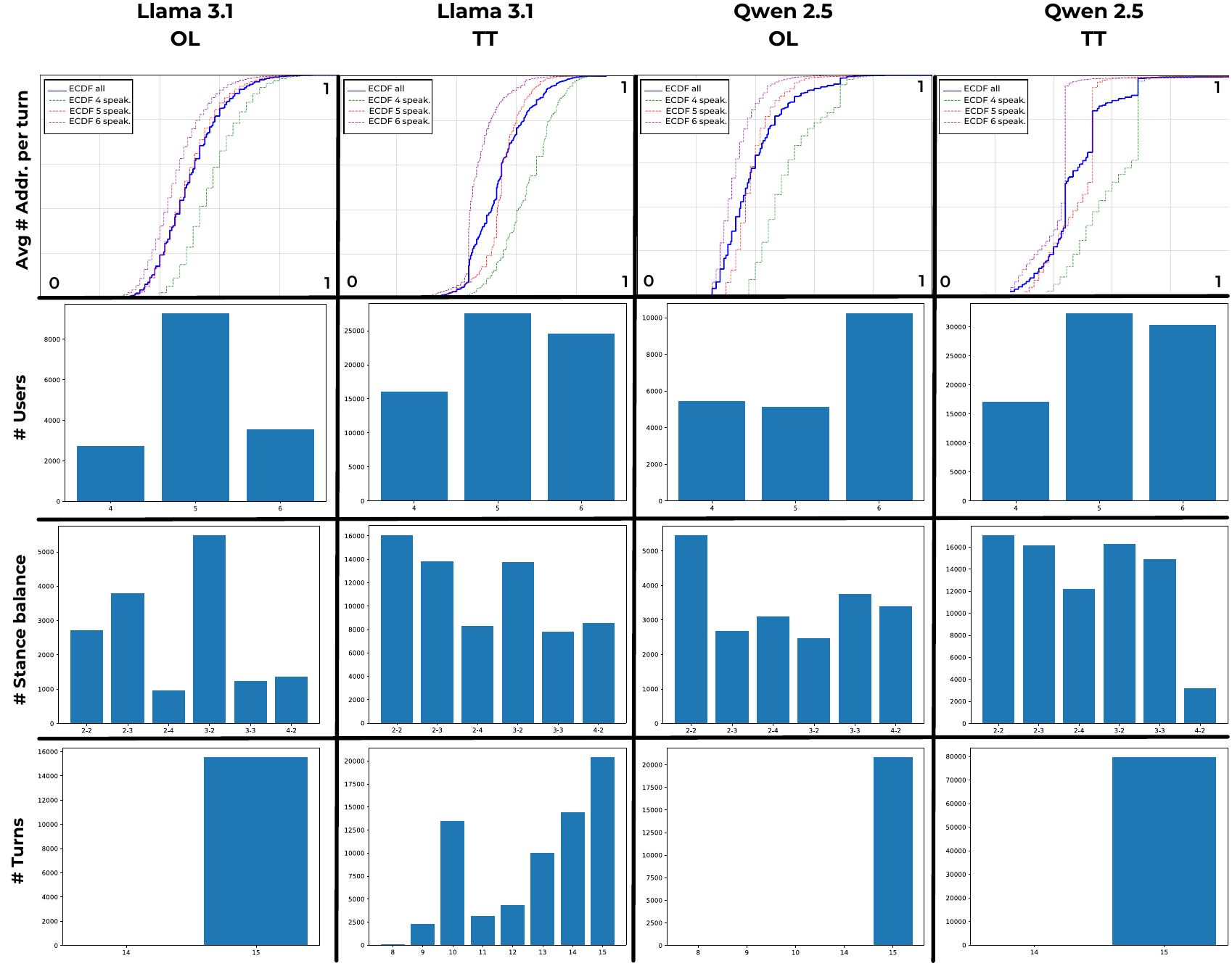}
    \caption{General statistics of the resulting WMPC for Llama3.1 and Qwen2.5 on both generation strategies. The statistics reported are (from the top): \textsc{(i.)} average number of addressees per turn, \textsc{(ii.)} number of users, \textsc{(iii.)} stance assignment, \textsc{(iv.)} number of turns.}
    \label{fig:general-stats}
\end{figure*}

In Figure \ref{fig:general-stats} we report the general statistics of the synthetic WMPCs that satisfy the constraints according to Section \ref{subsec:result_constr}. We report for each model-strategy combination: \textsc{(i.)} the average number of addressees per turn; \textsc{(ii.)} the number of users; \textsc{(iii.)} the stance assignment; \textsc{(iv.)} the number of turns.

Most of the combinations satisfy the strict requirements of exactly $15$ messages. Only Llama3.1-TT tends to generate a lot of examples of shorter conversations. 

\subsection{Similarity scores between conversations.}\label{app:similarity_scores}
 We provide a detailed explanation of how similarity scores were computed between WMPCs on the same topic.

\textit{Repetition Rate} is computed within each topic-based cluster by measuring repetitions  across all WMPCs in the cluster. The values are then averaged across clusters to assess the overall linguistic diversity.

\textit{String Similarity} and \textit{Semantic Coherence} are computed following an all-vs-all comparison strategy across every turn in all possible conversation pairs within the same topic.

Since Repetition Rate already reflects similarity across entire conversations, String Similarity is designed to penalize (with higher scores) pairs of conversations that contain also few highly similar turns, so promoting turn-level diversity.
Specifically, given two conversations with $a$ and $b$ turns respectively, we compute string similarity scores for all $a \times b$ turn pairs.
We then take the average of the top $5$ scores as the String Similarity for that conversation pair.

In contrast, Semantic Coherence aims to measure how consistently coherent the conversations during all the turns. So, for each of the 
$a$ turns in the first conversation, we keep the highest cosine similarity with any of the $b$ turns in the second conversation, and vice versa. The final Semantic Coherence for that conversation pair score is the average of these $a+b$ scores.
For both String Similarity and Semantic Coherence, the final topic-level value corresponds to the average across all conversation pair-scores of that topic.

\begin{figure*}
    \centering
    \includegraphics[width=0.95\linewidth]{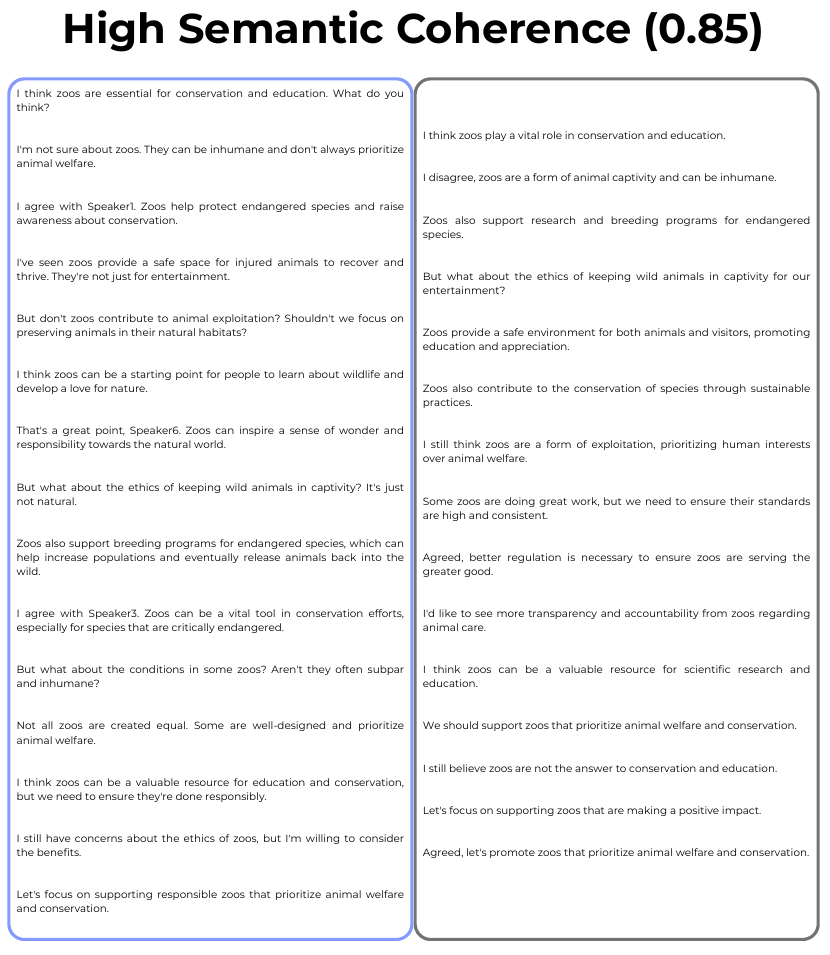}
    \caption{Conversations with highest Semantic Coherence from Llama3.1-OL.}
    \label{fig:high_semsim}
\end{figure*}

\begin{figure*}
    \centering
    \includegraphics[width=0.95\linewidth]{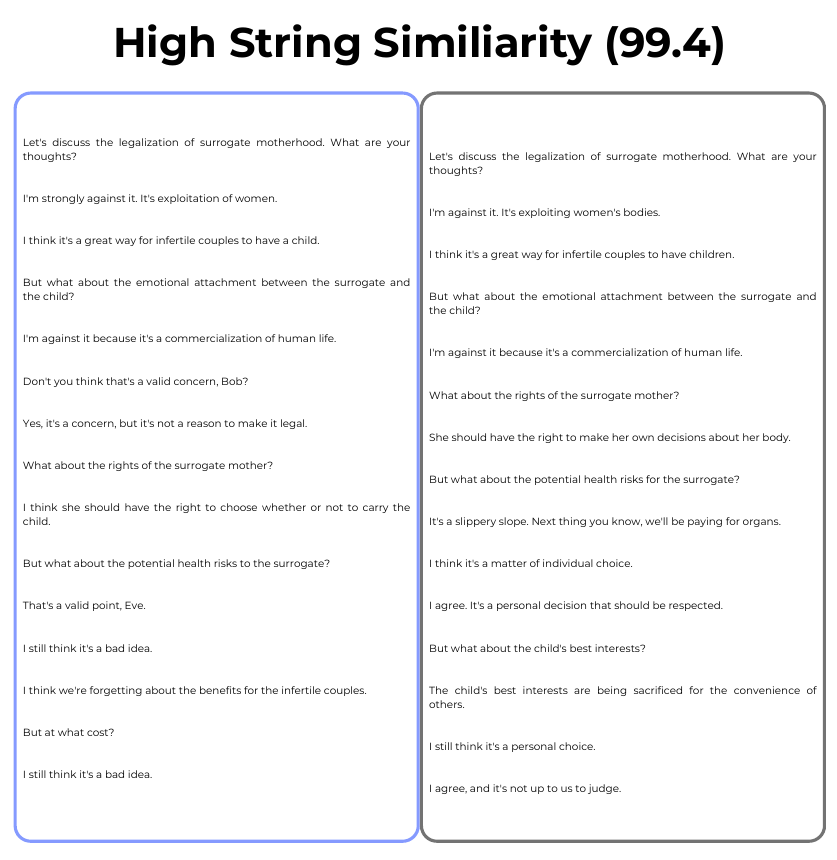}
    \caption{Conversations with highest String Similarity from Llama3.1-OL.}
    \label{fig:high_strsim}
\end{figure*}

In Figure \ref{fig:high_semsim} and Figure \ref{fig:high_strsim} we report  two pairs of conversations  generated with Llama3.1, One-Long strategy (i.e., the combination with highest Repetition Rate). We report in Figure \ref{fig:high_semsim} the pair of conversations with highest Semantic Coherence (equal to $0.85$) and in Figure \ref{fig:high_strsim} the pair of conversations with highest String Similarity (equal to $99.4$) from the pool of WMPCs generated by the LLama3.1-OL combination. We observe that, despite the very high similarity scores, the two conversations are still clearly different, ensuring a good variability in conversations generated with the same prompt.

\subsection{Extended Results from Global Structure Analysis}\label{appendix:global_extended}
\begin{figure*}
    \centering
    \includegraphics[width=0.95\linewidth]{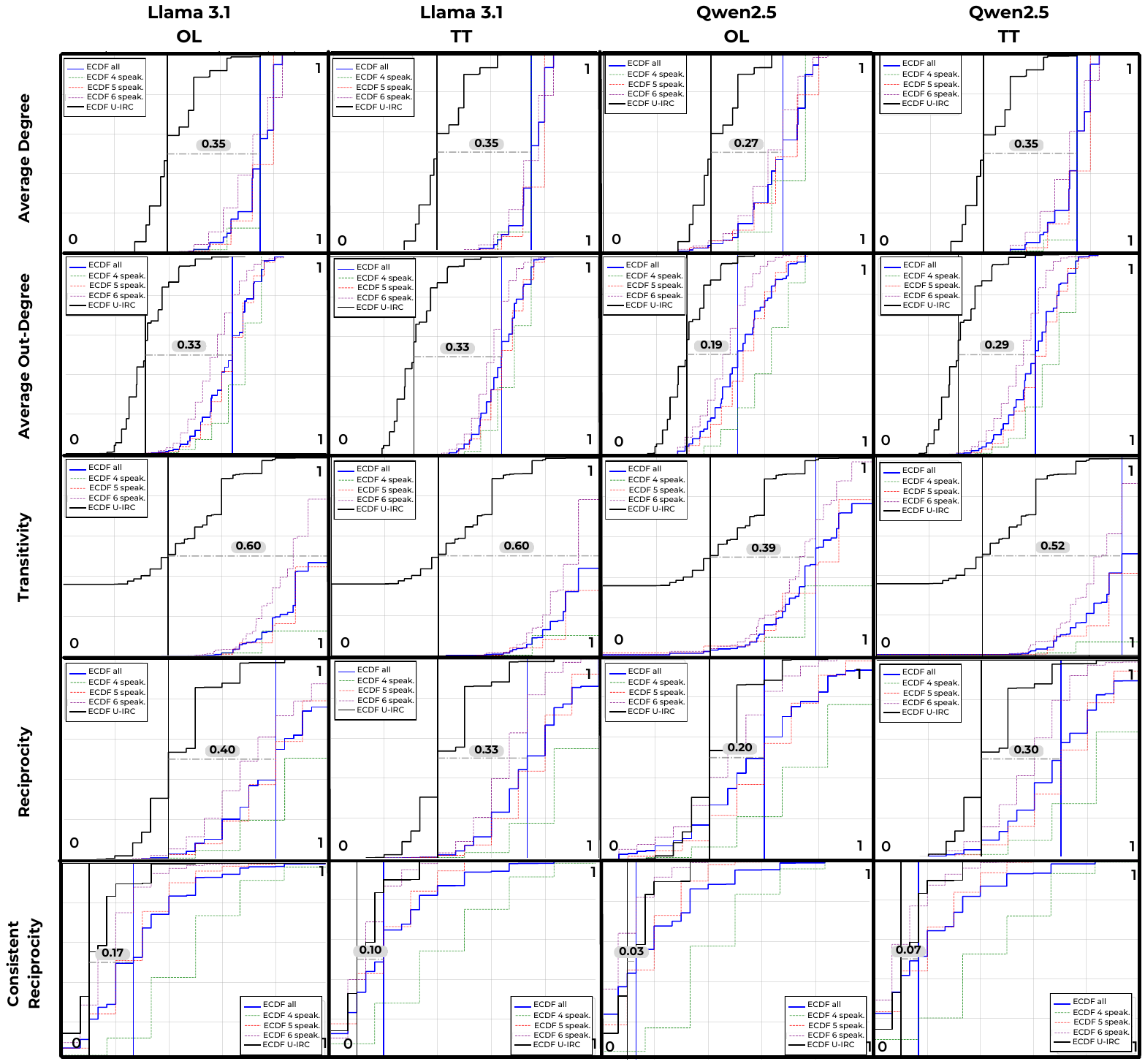}
    \caption{Empirical Cumulative Density Function (ECDF) of structural analysis on the synthetic WMPCs from Llama3.1 and Qwen2.5, with both generation strategies, i.e. One-Long and Turn-by-Turn generation. The statistics reported are (top to bottom): \textsc{(i.)} Average Degree Centrality, \textsc{(ii.)} Average Out-Going Degree, \textsc{(iii.)} Transitivity, \textsc{(iv.)} Reciprocity, \textsc{(v.)} Consistent Reciprocity. Average Degree Centrality and Average Out-Going Degree are normalized. In this way, all values on the vertical axis (density) and on the horizontal axis (value of the metric) are included between $0$ and $1$.}
    \label{fig:all-global}
\end{figure*}

In Figure \ref{fig:all-global} we report all the plots related to the $5$ structural metrics presented in Section \ref{subsec:glob}, for all the model-strategy combinations (where the models are Llama3.1 and Qwen2.5). All the plots show similar shapes, and all the combinations present a more complex structure than WMPCs in the UbuntuIRC dataset \cite{ouchi-tsuboi-2016-addressee}.

\section{Technical Report}\label{appendix:technical_report}

All the experiments have been run on Ampere A40 GPUs, which present 48GB of VRAM. We used the vLLM library \cite{kwon2023efficient} for speeding up the inference time, namely the version 0.6.6. In Table \ref{tab:models} we report the links to the models  and the repositories we used. As hyperparameters, we use temperature $0.7$, mixed top p and top k decoding with $p= 0.9$ and $k=40$. 

In Table \ref{tab:comp_time} we report the computational time to generate the full amount of $102\,600$ synthetic WMPCs for each model-generation strategy combination. OLMo2 is the model requiring more time (close to Ministral in One-Long strategy and definitely higher in Turn-by-Turn). In terms of strategies, as expected, the Turn-by-Turn is the one requiring more time, going from $\times4.05$ times more in Ministral to $\times7.82$ times more in Qwen2.5.

In order to compute the structural metrics, we used the tools from the \texttt{NetworkX}\footnote{\url{https://networkx.org/documentation/stable/}} library \cite{osti_960616}. Instead, for computing the Krippendorf-alpha we used the implementation from \citet{castro-2017-fast-krippendorff} for interval data. For the Spearman's correlation, we used the SciPy library \cite{2020SciPy-NMeth}. For the human evaluation, we used Argilla\footnote{\url{https://github.com/argilla-io/argilla/}} for creating the annotation platform and we used the User Interface provided on HuggingFace spaces.\footnote{\url{https://huggingface.co/argilla}} For the LLM-as-a-judge evaluation, we employed the OpenAI API for reasoning models.\footnote{\url{https://platform.openai.com/docs/guides/text-generation}} We also used the official guide to perform prompt refinement.\footnote{\url{https://platform.openai.com/docs/guides/prompt-generation}} Our results can be rerun by paying less than $\$20.00$. We used both ChatGPT\footnote{\url{https://openai.com/index/chatgpt/}} and Copilot\footnote{\url{https://github.com/features/copilot}} for help in the coding process.

\begin{table*}[ht!]
    \centering
    \begin{tabular}{|l|r|}
        \hline
        Model & Repository \\
        \hline
        Llama-3.1-8B & \url{https://huggingface.co/meta-llama/Llama-3.1-8B} \\
        Qwen2.5-7B-Instruct & \url{https://huggingface.co/Qwen/Qwen2.5-7B-Instruct}\\
        Ministral-8B-Instruct & \url{https://huggingface.co/mistralai/Ministral-8B-Instruct-2410} \\
        OLMo2-7B-Instruct & \url{https://huggingface.co/allenai/OLMo-2-1124-7B-Instruct}\\
        \hline
    \end{tabular}

        \begin{tabular}{|l|r|}
        \hline
        Model & Context length \\
        \hline
        Llama-3.1-8B & $128$k \\
        Qwen2.5-7B-Instruct & $131$k\\
        Ministral-8B-Instruct & $131$k \\
        OLMo2-7B-Instruct & $4$k\\
        \hline
    \end{tabular}
    
    \caption{Upper - repositories of the model used for generating the synthetic WMPCs. Lower - context length of each model.}
    \label{tab:models}
\end{table*}

 \begin{table}[ht]
    \centering
    \small
    {
    \begin{tabular}{|l|c|c|c|c|}
    \hline
  \textbf{Model} & \textbf{Llama3.1} & \textbf{Qwen2.5} & \textbf{Ministr.} & \textbf{OLMo2} \\ \hline
        \textit{OL}  & $1485$  & $1128$ & $2553$ & $2446$\\ 
        \textit{TT}  & $11540$  & $8824$ & $10335$ & $16580$ \\
        \hline
    \end{tabular}
    }
    \caption{Computational time for each model-strategy combination (in minutes).}
    \label{tab:comp_time}
\end{table}

\section{Human and LLM-as-a-judge Agreement} \label{appendix:agreement}

As reported in Section \ref{subsec:qualitative_res}, we compute Inter-Annotator Agreement (Krippendorf's alpha and  Spearman's correlation) on a batch of $96$ WMPCs between human  annotators and between each human annotator and the LLM (results in Table \ref{tab:agreement_human-llm}). Then, we compute Krippendorf's alpha among the human annotators and the LLM. For the score with highest Krippendorf's alpha, i.e., \textit{Speaker Stance Consistency}, we employ LLM-as-a-judge on large scale. For completeness, we run LLM-as-a-judge also on the other stance-based dimension, i.e., \textit{Speaker Stance Evolution}, where the correlation is still highly statistically significant, but with lower inter-annotator agreement (still positive).

The results highlight the degree of subjectivity of some evaluation scores and provide insights into the generally low inter-annotator agreement (IAA) between LLM-as-a-judge and human annotators. Notably, all dimensions, except \textit{Naturalness}, show statistically significant correlations between the two human annotators. Stance-based dimensions exhibit the highest IAA, followed by addressee-based and then stylistic ones. This pattern could be due to the annotators' differing backgrounds, one in philosophy, the other in computer engineering, which likely influenced their assessments of conversational quality (\textit{Naturalness}) and argumentative strength (\textit{Argumentability}). %
Agreement between LLM-as-a-judge and human annotators follows a similar trend, except for Addressee Preciseness and Stance Evolution, where agreement is markedly lower, suggesting that LLMs struggle particularly with assessing whether the set of addressees of a turn is too  generic and whether the stance evolution is plausible (with respect to humans).

\begin{table*}[ht!]
    \centering
    \small
    \begin{tabular}{|l|c|c|c|c|c|c|c|}
    \hline
\textbf{Annotators} &  \multicolumn{2}{c|}{\textbf{H1-H2}} & \multicolumn{2}{c|}{\textbf{H1-O3}} & \multicolumn{2}{c|}{\textbf{H2-O3}} & \textbf{H1-H2-O3} \\ \hline
\textit{Coefficients} &  \textit{Kr.} & \textit{Sp.} & \textit{Kr.} & \textit{Sp.} & \textit{Kr.} & \textit{Sp.} & \textit{Kr.}  \\ \hline
        Naturalness & $0.00$ & $0.09$ & $-0.24$ & $0.05$ & $0.16$ & $0.26^*$ & $-0.01$\\
        Argumentability  & $0.23$ & $0.36^{**}$ & $0.14$ & $0.22^*$ & $0.29$ & $0.31^*$ & $0.22$\\
        Addressee Correctness & $0.37$ & $0.37^{**}$ & $0.18$ & $0.30^*$ & $0.30$ & $0.42^{**}$ & $0.30$ \\
        Addressee Preciseness &  $0.41$ & $0.47^{**}$ &  $-0.08$ & $-0.12$ & $-0.04$ & $0.01$ &  $0.10$\\ \hline
        Stance Consistency  & $0.81$ & $0.77^{**}$ & $0.78$ & $0.76^{**}$ & $0.81$ & $0.78^{**}$ & $0.80$ \\
        Stance Evolution &  $0.49$ & $0.57^{**}$  &  $0.29$ & $0.25^*$ & $0.27$ & $0.42^{**}$ & $0.23$\\ \hline
    \end{tabular}
    \caption{Inter-annotator agreement (Krippendorf's alpha and Spearman's correlation)  between LLM as a judge (O3) and Human experts (H1 and H2). For  Spearman's correlation, (*) highlights that the correlation is statistically significant ($p < 0.05$). Instead, (**) corresponds to a correlation that is highly statistically significant ($p < 0.001$).}
    \label{tab:agreement_human-llm}
\end{table*}

\section{Guidelines for Human Evaluation}\label{appendix:guidelines}

\begin{figure*}
    \centering
    \includegraphics[width=0.8\linewidth]{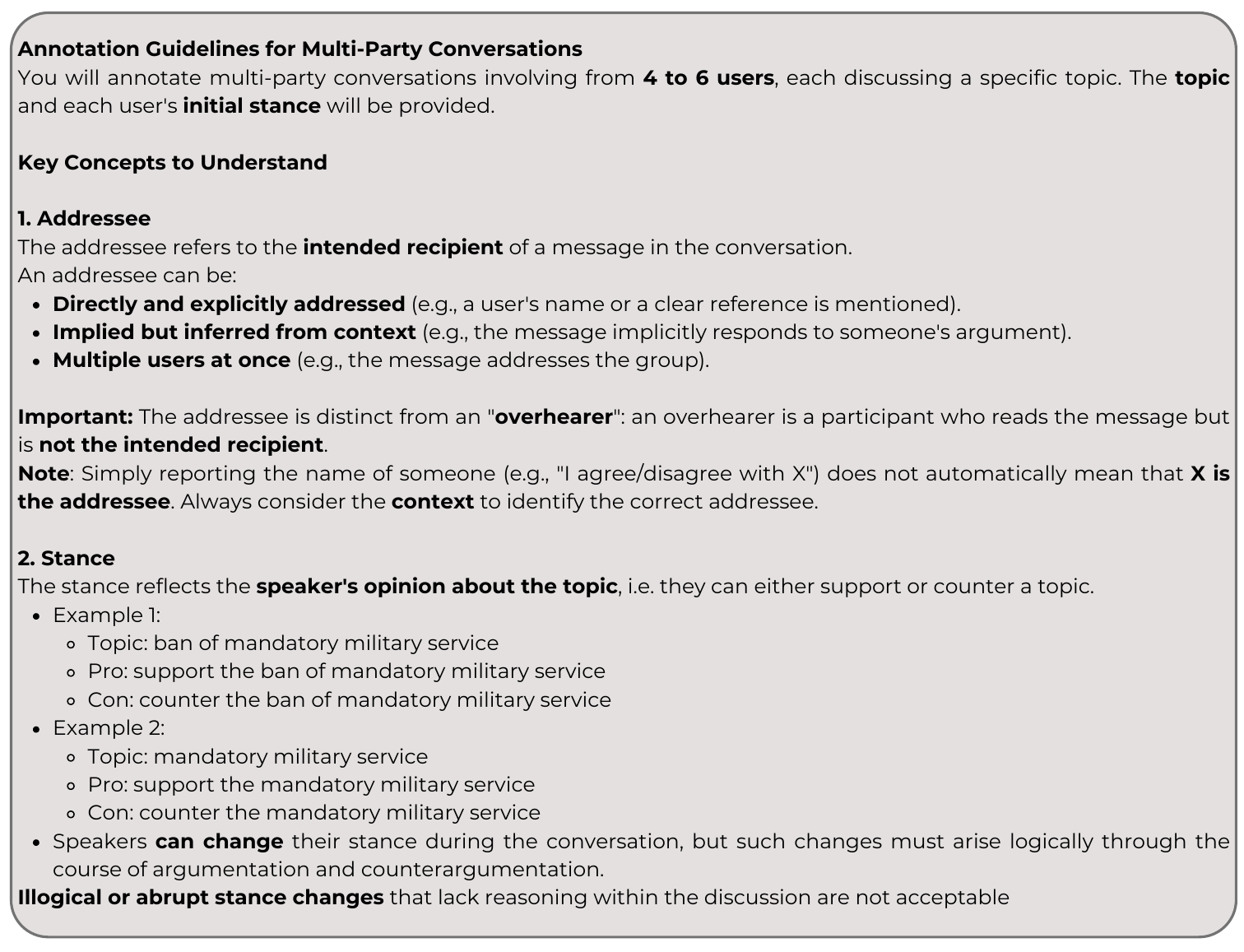}
    \caption{Overview of guidelines for human evaluation}
    \label{fig:human_eval_1}
\end{figure*}

\begin{figure*}
    \centering
    \includegraphics[width=0.8\linewidth]{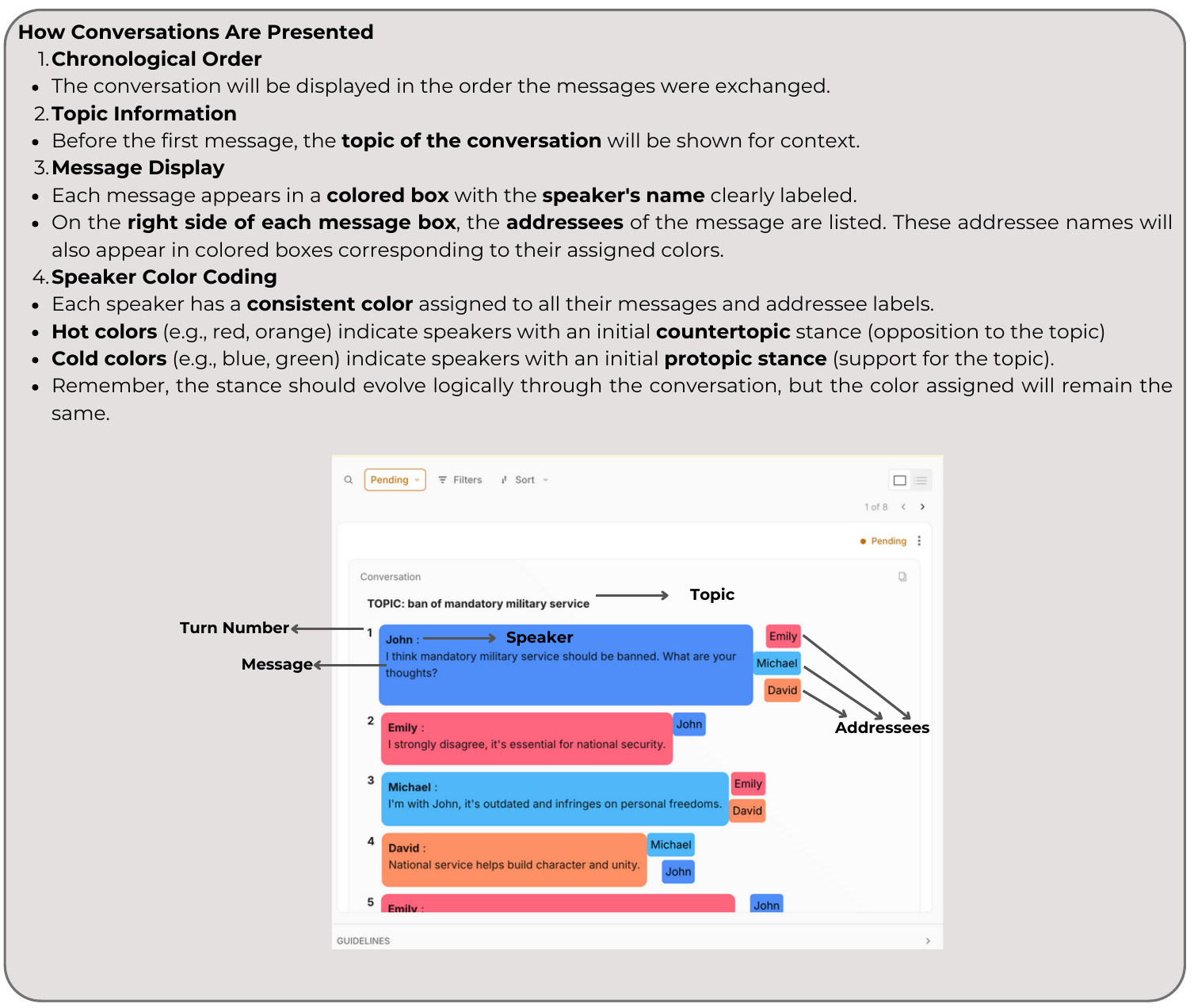}
    \caption{Platform description from human evaluation guidelines}
    \label{fig:human_eval_2}
\end{figure*}

\begin{figure*}
    \centering
    \includegraphics[width=0.8\linewidth]{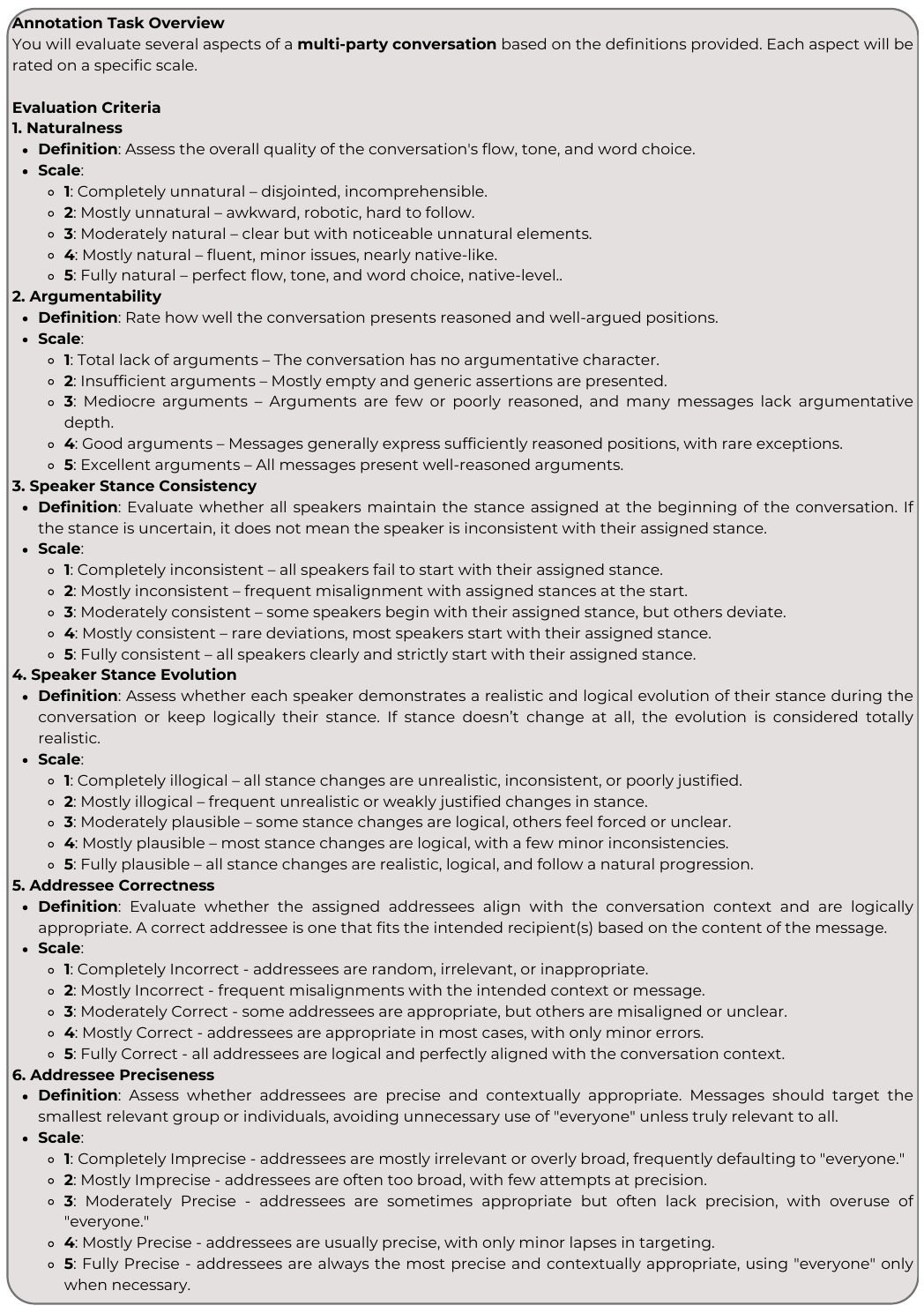}
    \caption{Description of the scores from the human evaluation guidelines}
    \label{fig:human_eval_3}
\end{figure*}

For human annotation, we designed a careful documentation to be used as  guidelines. We report in Figure \ref{fig:human_eval_1} the introduction, in Figure \ref{fig:human_eval_2} the Platform Description (with a screenshot of the view), in Figure \ref{fig:human_eval_3} the Scores Description.
One annotator performed his task as part of an internship, while the second annotator is regularly employed at the authors' institution. The effort required, on average, $2.5$ hours for a batch of $32$ items.

\end{document}